\documentclass[preprints,article,accept,moreauthors,pdftex,10pt,a4paper,color=usenames, dvipsnames]{mdpi}

\firstpage{1}
\usepackage{amsmath}
\usepackage{graphicx}
\usepackage{subcaption}
\usepackage{caption}
\usepackage{pseudocode}
\usepackage[official]{eurosym}
\usepackage{url}
\def\UrlBreaks{\do\/\do-}
\usepackage{tikz}
\usetikzlibrary{patterns}
\usetikzlibrary{shapes}
\usepackage[vlined,ruled]{algorithm2e}
\usepackage[utf8]{inputenc}

\history{Received: date; Accepted: date; Published: date}

\Title{From a Point Cloud to a Simulation Model: Bayesian Segmentation and Entropy based Uncertainty Estimation for 3D Modelling}

\Author{Christina Petschnigg $^{1,}$*\orcidA{}, Markus Spitzner$^{1}$, Lucas Weitzendorf$^{1}$ and J\"urgen Pilz $^{2}$\orcidB{}}

\AuthorNames{Christina Petschnigg, Markus Spitzner, Lucas Weitzendorf and J\"urgen Pilz}

\address{%
$^{1}$ \quad Department of Factory Planning, BMW Group, Knorrstra\ss e 147, 80788 Munich; Firstname.Lastname@bmw.de\\
$^{2}$ \quad Department of Statistics, Alpen-Adria-University Klagenfurt, Universit\"atsstra\ss e 65-67, 9020 Klagenfurt; Juergen.Pilz@aau.at}

\corres{Correspondence: Christina.Petschnigg@bmw.de}

\abstract{
The 3D modelling of indoor environments and the generation of process simulations play an important role in factory and assembly planning. In brownfield planning cases existing data are often outdated and incomplete especially for older plants, which were mostly planned in 2D. Thus, current environment models cannot be generated directly on the basis of existing data and a holistic approach on how to build such a factory model in a highly automated fashion is mostly non-existent. Major steps in generating an environment model in a production plant include data collection and pre-processing, object identification as well as pose estimation. In this work, we elaborate a methodical workflow, which starts with the digitalization of large-scale indoor environments and ends with the generation of a static environment or simulation model. The object identification step is realized using a Bayesian neural network capable of point cloud segmentation. We elaborate how the information on network uncertainty generated by a Bayesian segmentation framework can be used in order to build up a more accurate environment model. The steps of data collection and point cloud segmentation as well as the resulting model accuracy are evaluated on a real-world data set collected at the assembly line of a large-scale automotive production plant. The segmentation network is further evaluated on the publicly available Stanford Large-Scale 3D Indoor Spaces data set. The Bayesian segmentation network clearly surpasses the performance of the frequentist baseline and allows us to increase the accuracy of the model placement in a simulation scene considerably.}


\begin{document}



\section{Introduction}
\label{sec:introduction}
The Association of German Engineers (VDI) defines factory planning as "systematic, objective-oriented process for planning a factory, structured into a sequence of phases, each of which is dependent on the preceding phase, and makes use of particular methods and tools, and extending from the setting of objectives to the start of production"~\citep{verband2011vdi}. Four major types of factory planning scenarios are distinguished, namely development or greenfield planning, replanning or brownfield planning, dismantling and revitalization. The most common planning scenario is brownfield planning followed by greenfield planning.\\
Greenfield planning refers to planning a completely new plant with a new building, production line and processes. The generation of three-dimensional~(3D) simulations and the introduction of changes to the virtual production system are relatively easy as new plans fulfil current requirements of factory planning including 3D building plans, 3D layouts and computer aided design~(CAD) models of the facility, machinery and products. Further, new technologies and the need for digitalization are taken into account during planning. For instance, cameras and markers are added to the plans for later digitalization. Due to the availability of a digital model of the production system, planners are able to detect planning mistakes mostly in the virtual world before the actual implementation. Brownfield planning in contrast refers to the renovation or reorganization of an existing plant. This task is much more difficult compared to greenfield planning as existing layouts are often outdated and only available in two-dimensional~(2D) form. These plans hide substantial depth information, which is especially important when tasks are related to the introduction of new machinery or process steps. Due to the enormous area of assembly plants, which easily reaches up to several thousands of square metres, various problems arise. The determination of the as-is state of these plants is difficult, as layouts are often outdated and do not reflect all the renovations of the facility and the production line. The spatial requirements for a new planning case can hardly be determined by manual measurements due to the vast area of the plants. Further, certain technological constraints have to be accepted or modified with great effort. For instance, in older plants cabling, out-lets and interfaces for newer digitalization technologies are limited or even non-existent, which makes them difficult to digitalize. As the share of virtual planning is small compared to analogue planning many planning mistakes can only be detected during implementation, which is costly. The major differences of greenfield and brownfield planning are summarized in Table~\ref{tab:greenfield-vs-brownfield}.

\begin{table}[h] 
\centering
\caption{Major differences between greenfield and brownfield planning.}
\begin{tabular}{c|c}
\hline
\textbf{Greenfield Planning} & \textbf{Brownfield Planning}\\
\hline
Planning of a new plant & Renovation of an existing plant\\
3D building plans & Often only 2D building plans\\
3D CAD models of machines and tools & 2D construction drawings\\
Up-to-date data & Often outdated data\\
Decide for technologies to be used & Live with the technologies given\\
Regular digitalization taken into account & Difficult to digitalize\\
Learn from mistakes in simulations & Live with mistakes of direct implementation\\
\hline
\end{tabular}
\label{tab:greenfield-vs-brownfield}
\end{table}

In order to cope with the challenges arising in brownfield planning scenarios virtual planning undoubtedly provides great potential in terms of planning efficiency and accuracy. It has many advantages over analogue planning, e.g. virtual commissioning and virtual correction of mistakes. The highest economic benefit is the detection and correction of planning mistakes in early project phases -- ideally before their actual implementation~\citep{kuhn2006digital}. Further, factory planners may inspect overseas sites in a purely virtual environment to discuss modifications or other planning related tasks. A virtual 3D copy of the real plant can be transferred to virtual reality supported environments or simpler simulation engines, which usually provide the possibility of multi-user meetings. In the long run this saves a considerable amount of travel time and cost. Ideally, solely 3D models and corresponding simulations of factory buildings, inventory assets, product information and process steps are the basis for implementing assembly reorganizations including the introduction of new or the modification of existing processes.\\
In this work, we focus on the brownfield planning case and propose a holistic and methodical workflow based on~\citep{petschnigg2020point} for the automated generation of an environment model in a simulation engine starting from a raw point cloud of an industrial manufacturing plant. Initially, a data collection methodology based on 3D laser scanning and photogrammetry is proposed. This enables us to collect a data set at a German automotive OEM in order to thoroughly test our remaining workflow steps. The digitalization process yields a 3D point cloud of the factory environment. Objects within this point cloud are identified using a segmentation approach. To this end, we apply the Bayesian neural network (BNN) proposed in our previous work~\citep{petschnigg2020uncertainty}. The Bayesian architecture allows us to quantify the network's uncertainty in its predictions. Among others we use an entropy based uncertainty measure in order to identify uncertain predictions and improve the segmentation quality. This uncertainty definition enables us to distinguish between overall, data related and model related uncertainty. Such a differentiation is beneficial, as it can be determined to what extent uncertainty can be reduced by model refinement and to what extent uncertainty is inherent to the data. The segmentation output is used to determine the number of objects per scene as well as their pose relative to a predefined zero point. These steps are carried out by the application of clustering and point cloud registration algorithms. In summary, the contributions of this paper are:
\begin{itemize}
\item Workflow: We describe a comprehensive and methodical workflow starting with the digitalization of large-scale industrial environments using laser scans and photogrammetry and ending with the generation of a static environment model.
\item Experiment: We evaluate the quality of factory digitalization in terms of the accuracy, completeness and point density of the resulting point cloud. The efficiency of the Bayesian segmentation network and the quality of uncertainty information is briefly touched. An extensive evaluation can be found in~\citep{petschnigg2020uncertainty}. In this paper we further elaborate on the accuracy of the final model placement.
\item Potential: We provide an estimation of the economic potential of automated factory digitalization as well as simulation model generation for a number of exemplary production plants.
\end{itemize}
The remainder of this paper is structured as follows. In Section~\ref{sec:literature-review} a thorough discussion of state-of-the-art literature in factory planning, processing of point clouds as well as Bayesian neural networks and uncertainty quantification is presented. Section~\ref{sec:workflow-description} focuses on the proposed workflow starting with data collection and pre-processing. The Bayesian segmentation network as well as uncertainty quantification are briefly discussed. Further, methodologies for pose estimation and simulation model generation are described. The evaluation of our workflow is presented in Section~\ref{sec:evaluation-and-analysis} including the description of the data sets used. The complete workflow is evaluated by collecting and processing our own automotive factory data set. Aside from the collected data set, the segmentation step is additionally evaluated on the publicly available Stanford Large-Scale 3D Indoor Spaces data set. Finally, Section~\ref{sec:discussion-and-conclusion} provides a discussion of the proposed workflow as well as its economic potential before concluding the paper.

\section{Literature Review}
\label{sec:literature-review}
In the sequel state-of-the-art research literature focusing on (digital) factory planning is discussed. Prior research in the fields of point cloud processing and pose estimation on the basis of 3D point clouds is presented. Further, key literature addressing Bayesian neural networks and uncertainty estimation is discussed. 

\subsection{Factory Planning and Digitalization}
Until the early 20$^{th}$ century, the manufacturing industry was apt to mass production mainly requiring robustness to interruptions. Since that time production systems had to change fundamentally due to shorter product lifecycles, decreasing lot sizes and progressive product customization~\citep{bauernhansl2014industrie}. The factory lifecycle accompanies the factory from its beginnings, i.e. the initial planning, to its end of life. It is mainly determined by the shortening product and process lifecycles. The factory lifecycle can be split into five distinct phases, namely, development, construction, start-up, operation and dismantling~\citep{schenk2009factory}.\\
According to the factory lifecycle, the factory planning process is part of the development as well as in the operation phase for greenfield and brownfield planning, respectively. The factory planning process itself can be divided into another seven distinct phases, which are described in more detail in the VDI guideline~5200~\citep{verband2011vdi}. The first two phases aim at the description of the objectives and the collection of preliminary planning data. During the third and fourth phase the planning concepts for structural elements are elaborated until a realizable status is reached. The final phases prepare for the realization. They supervise and support the factory start-up. All of these phases are accompanied by project management. Undoubtedly, the availability of current 3D factory data and simulation models facilitates the realization of any of these phases in greenfield as well as brownfield planning. However, in this work we focus on brownfield planning and the acquisition of current models mainly for the structural part, i.e. the third and fourth factory planning phase.\\
In order to arrive at a current digital model of a production site, there are several hurdles to overcome. The first and at the same time one of the most challenging problems is the collection of current data in the plants, each of which extends over a huge area. In recent years laser scanning~\citep{shellshear2015maximizing} and photogrammetry~\citep{luhmann2010close} have become popular measures for indoor digitalization, generating 3D point clouds of the environment. As soon as the necessary 3D data are available, they have to be pre-processed in order to be useful for factory planning. The pre-processing includes the fusion of data collected by different sensors and data cleaning. For the mere aim of a rough visualization of the environment, the availability of 3D point clouds is sufficient. Promising results for surface reconstruction on the basis of point clouds are presented in~\citep{huang2009consolidation,zhou2019detail}. In order to achieve photorealistic results, further manual efforts are needed. The generation of an environment or simulation model consisting of individual objects requires the localisation of these objects within the point cloud. This can be tackled by object detection networks that directly perform pose estimation or a segmentation framework, where object poses are estimated on the basis of the segmented point cloud. Both of these approaches are discussed in further detail in the next section. Another prerequisite for model generation is the availability of CAD models for the objects of interest, as many simulation tasks cannot be carried out solely on the basis of point clouds. Raw point clouds are subject to occlusions, which cause holes within the point cloud. Therefore, tasks like collision checking are potentially not reliable. Further, working with an unsegmented point cloud has the disadvantage of not being able to distinguish different objects automatically. They have to be selected manually, which makes their displacement or removal in a scene inconvenient.

\subsection{Learning on Point Clouds and Environment Modelling}
\label{subsec:learning-on-pc}
Learning frameworks on 3D point clouds have to cope with the unordered and irregular structure of the inputs. Some of these frameworks include~\citep{ravanbakhsh2016deep,qi2017pointnet,qi2017pointnet++}. There exist several strategies in order to transform point clouds to a more regular representation, namely voxelization or projection to the 2D space. In the former case a volumetric grid is created that represents the point cloud in a regular format. Different deep learning approaches, which use voxelized point clouds as an input, are discussed in~\citep{wu20153d,qi2016volumetric,zhou2018voxelnet}. In the latter case, the input point cloud is projected to the 2D space under varying viewports~\citep{chen2017multi,feng2018towards,yang2018pixor}. The transformation to a regular data structure has many advantages like the possibility to apply either 2D or 3D convolutions with their elaborate kernel optimizations. However, voxelization unnecessarily expands the data volume as empty areas in the point cloud are still represented by voxels. Further, dense areas of the point cloud are displayed in a lower resolution by the voxel grid, which entails an information loss~\citep{xie2020review} and truncation errors~\citep{qi2017pointnet}. The transformation of 3D point clouds to 2D images can cause the loss of structural information embedded in the higher dimensional space. Additionally, a high number of viewports need to be considered in order to represent the information encoded within a point cloud in a number of images~\citep{xie2020review}. Thus, we decide to work directly with the raw and unordered point cloud as an input for the segmentation network.\\
There exist several approaches to model the environment based on point clouds. Different objects in the point cloud need to be identified, which can be solved using various statistical or machine learning techniques. Segmentation is a way of assigning a distinct class label to each point in the point cloud, thus, allowing the differentiation of objects in the point set. Using the segmented point cloud, the 9DoF object poses, i.e. position, rotation and scale with respect to some predefined coordinate origin, can be calculated~\citep{petschnigg2020point}.\\
Aside from segmentation there is the possibility to perform direct object pose estimation in the point cloud. The Scan2CAD framework directly estimates the 6DoF CAD model alignment, i.e. position and rotation, of eight household objects on the basis of a voxelized point cloud and the respective CAD models~\citep{avetisyan2019scan2cad}. An extension of this framework, which is capable of 9DoF pose estimation on the same data set, is discussed in~\citep{avetisyan2019end}. Apart from deep neural networks a CAD model alignment framework based on global descriptors extracted by the Viewpoint Feature Histogram approach~\citep{rusu2010fast} is described in~\citep{aldoma2011cad}. Direct estimation of the CAD model alignment on the basis of point clouds can of course be used for environment modelling and simulation model generation. However, the prior segmentation of an input point cloud with a subsequent pose estimation step allows us to partition the point cloud into meaningful subsets of points, i.e. building, assembly, logistics and product related point sets. These subsets can be provided to the respective departments that only process the necessary parts of the point cloud, thus, reducing the storage and computational efforts considerably. For this reason, we rather apply the segmentation approach than direct pose estimation.

\subsection{Bayesian Neural Networks and Uncertainty Definition}
\label{subsec:BNN-Unc}
Bayesian neural networks (BNNs) in contrast to frequentist or classic neural networks take on the Bayesian interpretation of probability. They place a prior distribution over all the network parameters including weights and biases, instead of considering them as point estimates. Thus, a prior distribution needs to be defined for all the network parameters. The posterior distribution can be calculated using Bayes' theorem after each batch of training data. However, the direct application of Bayes' theorem is difficult as a generally intractable integral needs to be solved. There exist several techniques to approximate the posterior, e.g. variational inference~\citep{graves2011practical}, Markov Chain Monte Carlo approaches~\citep{hastings1970monte, brooks2011handbook,gelfand1990sampling}, Hamiltonian Monte Carlo algorithms~\citep{duane1987hybrid} and Integrated Nested Laplace Approximations~\citep{rue2009approximate}.\\
As already mentioned, BNNs allow the estimation of uncertainty in the network outputs, which can be realized using a number of different approaches. Many works distinguish between uncertainty inherent to the data and model related uncertainty. These are called aleatoric and epistemic uncertainty, respectively~\citep{der2009aleatory}. The total uncertainty in a network prediction, which is referred to as predictive uncertainty, can be interpreted as the sum of aleatoric and epistemic uncertainty. Such a split of uncertainty into data and model related uncertainty is useful as it allows us to grasp the amount of uncertainty that can be reduced by further model optimization and the amount of uncertainty, which is inherent to the data. One way of estimating uncertainty in a BNN is based on entropy~\citep{gal2017deep}. Precisely, predictive uncertainty $U_{pred}$ can be modelled as the entropy $\mathbb{H}$ inherent to the predictive network outputs, i.e. $U_{pred}~=~\mathbb{H}[y^{\star}\vert \underline{x}^{\star},\underline{w}]$, where $(\underline{x}^{\star},y^{\star})$ is an unseen input with its corresponding label and $\underline{w}$ corresponds to the network parameters. In practical applications the marginalization over the weight samples drawn from an approximate variational distribution yields $U_{pred}$,

\begin{equation}
U_{pred}~\approx~-\sum_{y^{\star}\in \{1,\dots,m\}}\left(\frac{1}{K}\sum_{k=1}^{K}p(y^{\star}\vert \underline{x}^{\star},~\underline{w}^{k})\right)\cdot log\left(\frac{1}{K}\sum_{k=1}^{K}p(y^{\star}\vert \underline{x}^{\star},~\underline{w}^{k})\right).
\label{eq:U-pred}
\end{equation}

In Equation~(\ref{eq:U-pred}), $p(y^{\star}\vert \underline{x}^{\star},~\underline{w}^{k})$ represents the predictive network output and $\underline{w}^{k}$ is the $k$-th weight sample of the variational distribution $q_{\underline{\varphi}}$, which is described in more detail in Section~\ref{subsubsec:workflow-bayesian-seg}. In total, $K\in\mathbb{N}$ weight samples are drawn from the variational distribution. The average entropy over all weight samples $\underline{w}^{k}$ corresponds to the aleatoric uncertainty,

\begin{equation}
U_{alea}~=~E[\mathbb{H}[y^{\star}\vert \underline{x}^{\star},\underline{w}^{k}]] ~\approx~ -\frac{1}{K}\sum_{k=1}^{K}\sum_{y^{\star}\in \{1,\dots,m\}}p(y^{\star}\vert \underline{x}^{\star},\underline{w}^{k})\cdot log(p(y^{\star}\vert \underline{x}^{\star},\underline{w}^{k})).
\label{eq:U-alea}
\end{equation}

Epistemic uncertainty $U_{ep}$ is calculated as the difference of predictive and aleatoric uncertainty, i.e. \mbox{$U_{ep}=U_{pred}-U_{alea}$}.\\
Further ways of uncertainty quantification include the variance of the predictive network outputs and the estimation of (95~\%-)credible intervals~\citep{steinbrener2020measuring}. In the former case the predictive variance is estimated empirically using the unbiased estimator for the variance. Any prediction for which the predictive variance exceeds a certain threshold, e.g. the mean variance plus two sigma, is considered uncertain. In the latter case, credible intervals are calculated for each class. If the credible interval of the predicted class overlaps with any other class' credible interval the prediction is considered uncertain. In~\citep{petschnigg2020uncertainty} all of these methods for uncertainty quantification are contrasted. Predictive as well as aleatoric uncertainty are able to effectively capture predictions that are uncertain and thus, potentially wrong. Excluding uncertain points according to these measures results in a considerably higher prediction accuracy while still maintaining an arbitrarily high ratio of points.

\section{Workflow Description}
\label{sec:workflow-description}
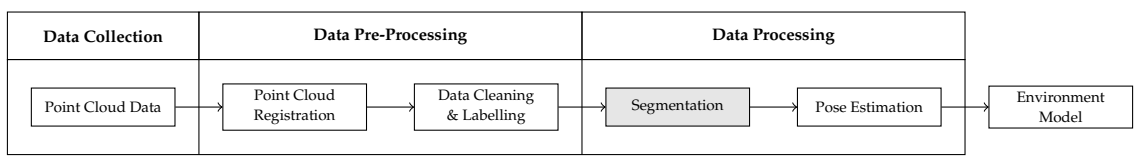
\begin{figure}[h]
\centering
        \begin{tikzpicture}[scale=0.63, every node/.style={scale=0.63}]

            \draw (1,0) rectangle (21,2); 
            \draw (5,0) -- (5,2); 
            \draw (13,0) -- (13,2); 
            
            \node (L1) at (3,1) [draw, minimum width=3cm, minimum height=0.8cm] {\small Point Cloud Data};
            \node (L2) at (7,1) [draw, minimum width=3cm, minimum height=0.8cm, align=center] {\small  Point Cloud\\ \small Registration};
            \node (L3) at (11,1) [draw, minimum width=3cm, minimum height=0.8cm, align=center] {\small  Data Cleaning\\ \small \& Labelling};
            \node (L4) at (15,1) [draw, minimum width=3cm, minimum height=0.8cm,  align=center, fill=gray!20] {\small  Segmentation};
            \node (L5) at (19,1) [draw, minimum width=3cm, minimum height=0.8cm,  align=center] {\small  Pose Estimation};
            \node (L6) at (23,1) [draw, minimum width=3cm, minimum height=0.8cm,  align=center] {\small  Environment\\ \small Model};
            
			\draw [->] (L1) edge (L2) (L2) edge (L3) (L3) edge (L4) (L4) edge (L5) (L5) edge (L6) (L6);         
            
            \node at (3,2.5) [draw, minimum width=4cm, minimum height=1cm] {\textbf{Data Collection}};
            \node at (9,2.5) [draw, minimum width=8cm, minimum height=1cm] {\textbf{Data Pre-Processing}};
            \node at (17,2.5) [draw, minimum width=8cm, minimum height=1cm] {\textbf{Data Processing}};
\end{tikzpicture}
\caption{Process diagram of generating an environment model on the basis of a point cloud. The segmentation step is highlighted as this step is explained in more detail in~\citep{petschnigg2020uncertainty}.}
\label{fig:process-sim-gen}
\end{figure}

The proposed workflow for generating a static simulation model consists of several steps. The major process steps include the methodical collection of suitable 2D and 3D data as well as data pre-processing. After pre-processing the data are further processed to yield an environment model. Following the reasoning of Section~\ref{subsec:learning-on-pc}, the generation of a static simulation model requires the segmentation of the input point cloud. We train the segmentation network in a Bayesian way, which enables us to estimate the network's uncertainty in its predictions. From the segmented point cloud object poses are estimated and the final factory model is created in a simulation environment. The workflow for the generation of a static simulation model is depicted in Figure~\ref{fig:process-sim-gen}. In the following the major components of this methodology are discussed. The segmentation step, which is depicted by the grey box in the figure is elaborated in more detail in~\citep{petschnigg2020uncertainty}.
\subsection{Data Collection}
\label{subsec:data-collection}
\begin{figure}[h]
\centering
\begin{subfigure}[c]{0.48\textwidth}
\includegraphics[width=0.97\textwidth]{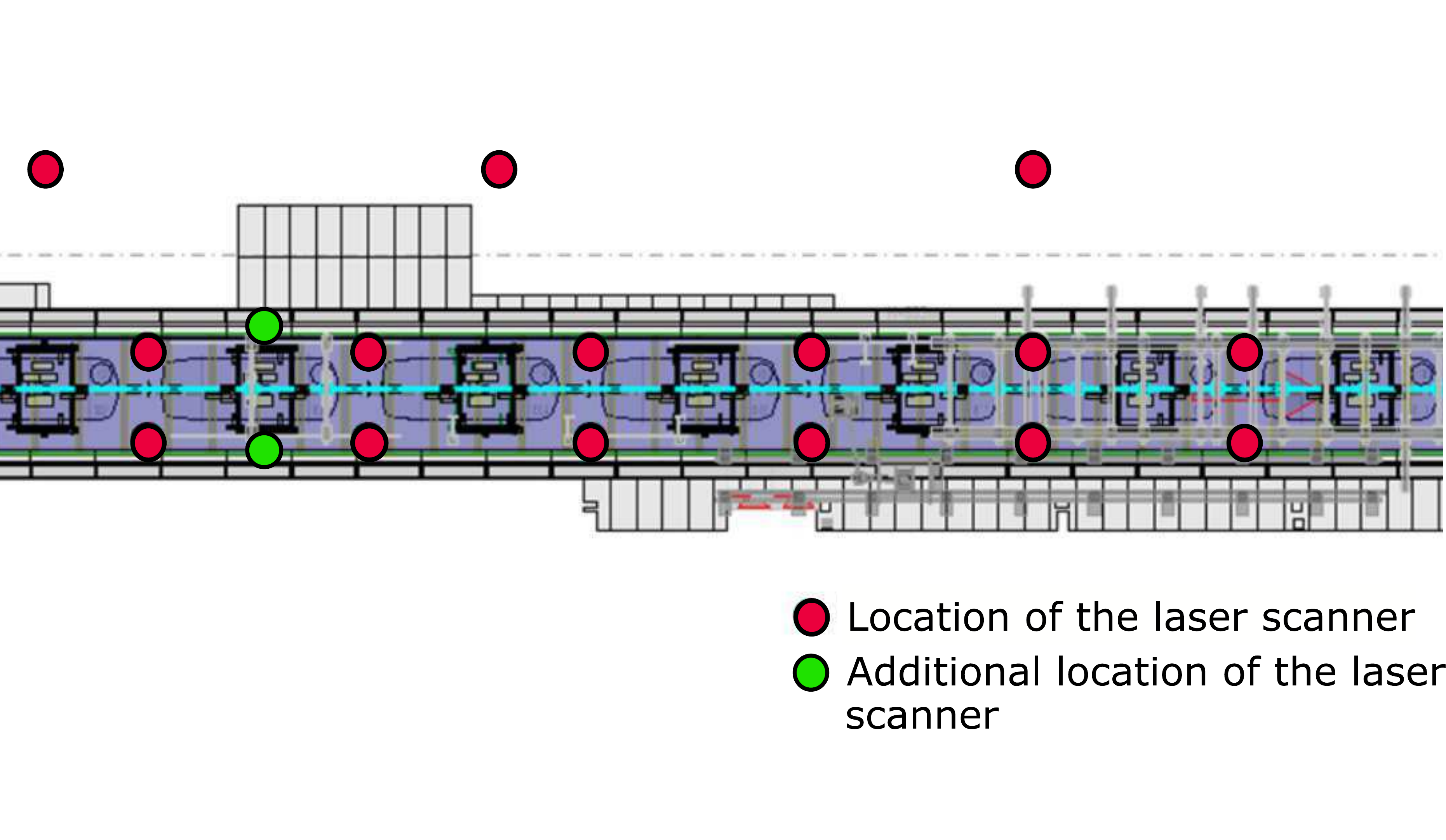}
\subcaption{}
\end{subfigure}
\begin{subfigure}[c]{0.48\textwidth}
\includegraphics[width=0.97\textwidth]{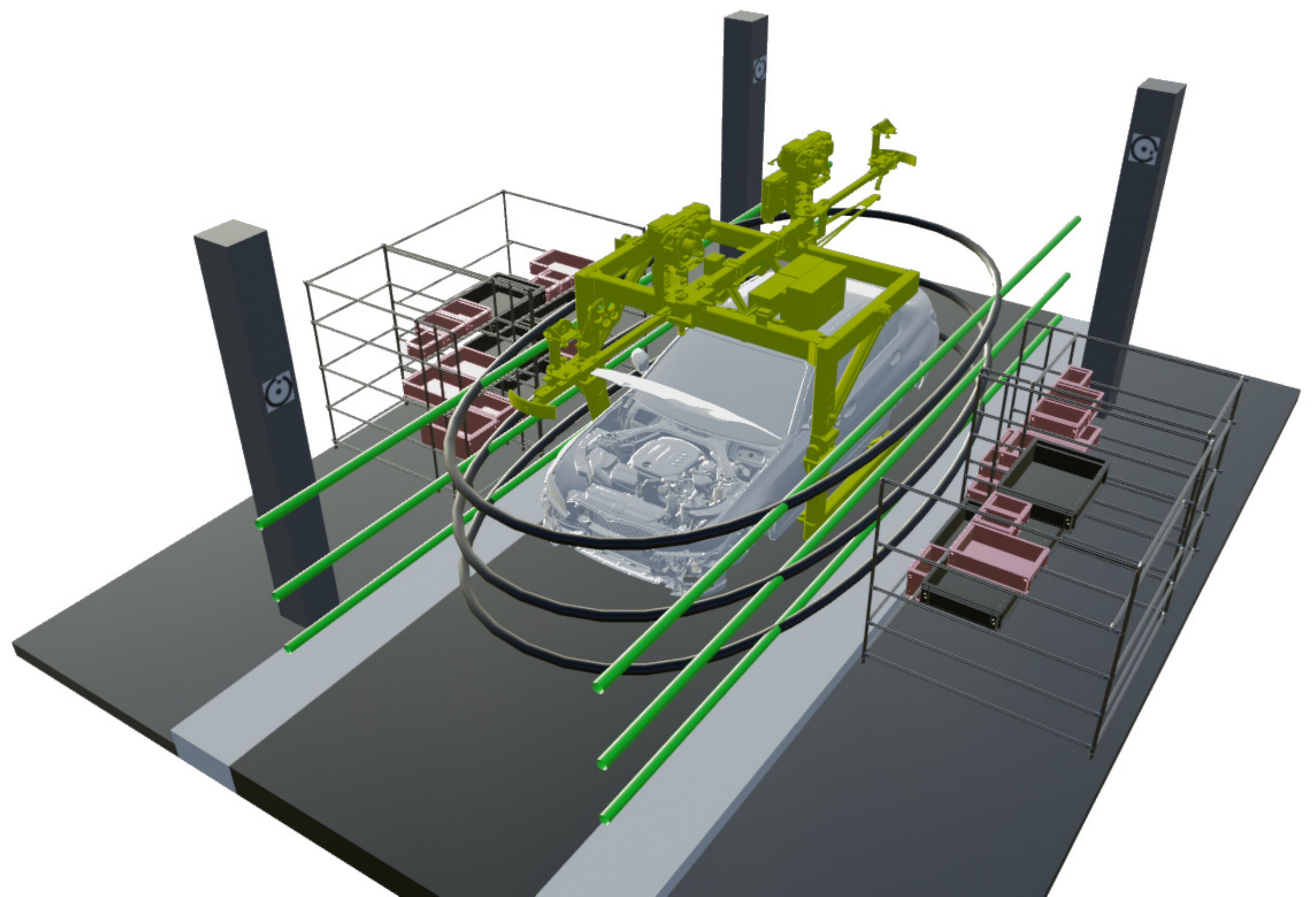}
\subcaption{}
\end{subfigure}
\caption{Locations of data collection. (\textbf{a}) Laser scanner positions for collecting four (red) and six (green) laser scans per tact. We collect four laser scans per tact as well as additional laser scans of the surrounding building to generate our data set. The six laser scan set up is used for evaluation purposes. \mbox{(\textbf{b}) Concept} of taking camera images.}
\label{fig:scanner-positions}
\end{figure}

Often high-quality data of the as-is state of production plants are not available. As already mentioned, brownfield planning scenarios often lack current factory data, sometimes not even 3D information is available. Further, not all renovation works are properly reflected in the layouts. Thus, over the years a plant layout can be up-to-date in some places and outdated in others, with 3D information being available only in some parts of the plant. Therefore, it is very important to find a way of methodical and periodically recurrent data collection to ensure the quality and timeliness of the plant layouts, CAD models and construction drawings of production and logistics equipment.\\
In recent years, laser scanning became increasingly popular for the purpose of indoor and outdoor digitalization~\citep{previtali2014flexible}. As a rule of thumb, stationary laser scanners produce a more accurate point cloud than mobile ones as they stand still during digitalization. Further, static laser scanners usually have a longer range than mobile ones. This insignificant in the plane but becomes important when measuring heights. However, stationary laser scanners produce point clouds with a high point density close to the device, which is decreasing over distance, while mobile laser scanners capture the environment with a lower but more homogeneous point density. Further, stationary scanners take a longer time to be set up and capture their surroundings with as few occlusions as possible~\citep{thomson2013mobile}. In order to reduce the number of stationary scans needed to generate a high-quality point cloud of the factory environment, additional sources of data can be added. For instance, mobile laser scanners or photogrammetry techniques or even both can be used to complement stationary laser scan point clouds. Thus, the number of holes in the resulting point cloud is reduced and the point density is increased. As not all applications of point clouds require the same resolution and accuracy, it is efficient to carry out base digitalization using mobile scanners while conducting stationary scans only where higher accuracy is really needed.\\
Before starting to digitalize any large-scale environment, it is advisable to place so-called registration targets in the facility in order to facilitate the fusion of different point clouds into one common coordinate system. After placing the targets, the actual digitalization process started. We use a static Faro Focus3D X 130HDR~\citep{FARO} laser scanner, a \mbox{Nikon D5500~\citep{Nikon}} camera with an 8~mm fish-eye lens as well as a \mbox{Sony Alpha 7R II~\citep{Sony}} with a 25~mm~wide-angle lens to capture the environment. Currently, an analysis of additional equipment is ongoing. For the following experiments, we digitalize several tacts of an automotive assembly plant by placing the static FARO scanner in different positions on the assembly line and next to it during the production free time. When choosing the scanner positions it is important to reduce object occlusions to a minimum, which results in a point cloud with a minimum number of holes. Figure~\ref{fig:scanner-positions}~(a) illustrates the scanner positions on a rough sketch of a factory layout for several tacts. In this layout four and six laser scans are marked per assembly tact. The four laser scans, which we use to digitalize the assembly area, are displayed in red and additional two laser scans are marked in green, which we use as a comparison. Further scans are generated outside of the assembly line in order to capture additional facility structures as well as production equipment.\\
In addition to the laser scans, we roughly take 500 camera images of each tact as well as several hundreds of the surrounding factory area. Prior to taking the pictures to generate the photogrammetric point cloud, both cameras with their respective lenses need to be calibrated in order to allow later image registration. We use a chequerboard calibration pattern and take images from different views in order to determine the camera parameters and calibrate the camera with the respective lens. Figure~\ref{fig:scanner-positions}~(b) shows the height at which the images were taken. The grey lines represent the height at which inward facing images were taken, i.e. images of the assembled vehicle. The green lines indicate the height at which outward facing images were taken, i.e. images of the lineside layout.
\subsection{Data Pre-Processing}
In order to generate a single point cloud displaying the factory building, all the different laser scan point clouds that have local coordinate systems need to be combined within a common global coordinate system. This process is called registration and is carried out using the dedicated scanner software. The registration targets are recognised automatically and further target points that are visible in several point clouds can be added manually in the case that registration cannot be completed using the targets alone. This provides a rough registration, which is improved with the iterative closest point~(ICP) algorithm~\citep{besl1992method}.\\
After the registration of the laser scan point clouds the image data is sorted according to the assembly tact and blurred images are removed, which is automated using a technique based on convolutions~\citep{pech2000diatom}. However, the removal of some of the images leaves us with a different number of images per tact, which can influence the resulting point cloud density, see Section~\ref{subsec:results-data-coll}. The information included in the remaining images has to be combined with the laser scan data. Due to the high number of reflecting objects within the environment like painted car bodies or metallic machines, the laser scan point cloud and the camera images are fused on image level. Synthetic images are created on the basis of the laser scan point cloud using the approach in~\citep{forkuo2004automatic}. These synthetic images are fused with the camera images using a structure from motion algorithm~\citep{westoby2012structure}. After the fused point cloud, which is made up of laser scans and camera images, is generated, it can still be very noisy. This is neither favourable for visualization nor for segmentation and object pose estimation. Thus, noise suppression and outlier detection algorithms are applied, which remove a number of noise points. In order to achieve a suitable degree of noise removal, the point cloud is further cleaned manually by the authors. In the course of this manual cleaning step the data set is labelled as well. For simulation model generation the point cloud needs to be segmented, which requires supervised training with labelled training data. It is important to identify objects that are relevant to factory planning before labelling the data. Relevant classes are mainly those objects that are immovable, fixed to the ground or part of the building as they need to be taken into account in the next renovation project.

\subsection{Point Cloud Segmentation}
As already discussed, the generated point cloud is segmented in order to identify the different objects in the scene. Therefore, we use the Bayesian segmentation framework, which we proposed in our previous work~\citep{petschnigg2020uncertainty}. In the sequel the mathematical notations and preliminary information on the segmentation step are detailed, followed by a brief description of the applied network.

\subsubsection{Notation and Preliminaries}
In Section~\ref{subsec:BNN-Unc} we already mentioned that BNNs place a probability distribution over all network parameters that is optimized during network training. In order to describe such a BNN model, the following notations are used throughout the remaining work. The network parameters include the weights and biases of all $d\in\mathbb{N}$ network layers, which are denoted by $\mathcal{W}$ and $\mathcal{B}$, respectively. The parameters of network layer $i$ are represented by $\mathcal{W}_{i}$ and $\mathcal{B}_{i},~i\in\{1,\dots,d\}$. The network inputs are given by $X=\{\underline{x}_{1},\dots,\underline{x}_{n}\}$ and the corresponding labels are depicted by $Y=\{y_{1},\dots,y_{n}\}$. The variables $n\in\mathbb{N}$ and $m\in\mathbb{N}$ represent the number of network inputs and the number of classes, respectively. Let further $p(\underline{w})$ denote the prior distribution of the network parameters. The a posteriori distribution of the network parameters after the evaluation of the training examples is given by $p(\underline{w}\vert Y,X)$ and can be calculated using the Bayes' theorem. The calculation of the posterior distribution is usually intractable as a result of the high dimensional sample space. Thus, in many cases it is not possible to find an analytical solution for the posterior distribution, however, several approximation techniques exist. The different posterior approximation methods mentioned in Section~\ref{subsec:BNN-Unc} are reviewed in our previous work~\citep{petschnigg2020uncertainty}. Due to efficiency reasons, we decide to use variational inference for posterior approximation. The main idea of variational inference is to approximate the intractable posterior distribution $p(\underline{w}\vert Y,X)$ with the so-called variational distribution $q_{\underline{\varphi}}(\underline{w})$, which is tractable. The variational distribution is parametric and the components of $\underline{\varphi}$ are referred to as the variational parameters. In order to approximate the posterior distribution, the distance between the posterior and the variational distribution has to be minimized. As an evaluation metric we use the Kullback-Leibler (KL) divergence, which is frequently used to determine the similarity of two distributions even though it is not a distance metric in the common sense. It neither satisfies the triangle inequality nor is it symmetric. As the KL-divergence cannot be minimized directly, the equivalent optimization problem of minimizing the negative log evidence lower bound is solved~\citep{bishop2006pattern}.\\
After the variational distribution $q_{\underline{\varphi}}(\underline{w})$ is optimized by processing the training samples $(X,~Y)$, the class label $y^{\star}$ of an unseen data point $\underline{x}^{\star}$ can be predicted. The posterior predictive distribution $p(y^{\star}\vert \underline{x}^{\star},Y,X)$ expresses the confidence in a label $y^{\star}$ after observing the input $\underline{x}^{\star}$. It is calculated using

\begin{equation}
p(y^{\star}\vert \underline{x}^{\star},Y,X)~=~\int p(y^{\star}\vert \underline{w},\underline{x}^{\star})p(\underline{w}\vert Y,X)~d\underline{w}.
\label{eq:posterior-predictive}
\end{equation}

In practice, the unknown posterior in Equation~(\ref{eq:posterior-predictive}) is replaced by the variational distribution and the integral is approximated by Monte Carlo integration, i.e.

\begin{equation}
p(y^{\star}\vert \underline{x}^{\star},Y,X)~\approx~\frac{1}{K}\sum_{k=1}^{K}f(\underline{x}^{\star},\underline{w}^{k}),~\mbox{with}~\underline{w}^{k}\underset{i.i.d.}{\sim}q_{\underline{\varphi}}(\underline{w}),
\end{equation}

where $f$ denotes a forward pass through the network and $K\in\mathbb{N}$ corresponds to the number of Monte Carlo weight samples  $\underline{w}^{k}$ drawn from the variational distribution. The predicted class $\hat{y}^{\star}$ is calculated using

\begin{equation}
\hat{y}^{\star}~=~\underset{y^{\star}\in\{1,\dots,m\}}{~\mbox{arg~max}~~}p(y^{\star}\vert \underline{x}^{\star},Y,X).
\end{equation}

\subsubsection{Bayesian Segmentation Network}
\label{subsubsec:workflow-bayesian-seg}
The network we use for the subsequent evaluations is described in more detail in our previous work~\citep{petschnigg2020uncertainty}. The variational distribution is based on the one proposed in~\citep{steinbrener2020measuring}. We define the random weights $\mathcal{W}_{i}:=(W_{i1},\dots,W_{ik_{i}})$ and biases $\mathcal{B}_{i}:=(B_{i1},\dots,B_{ik_{i}^{\prime}})$ of the $i$-th network layer, where $k_{i}\in\mathbb{N}$ denotes the number of weights and $k_{i}^{\prime}\in\mathbb{N}$ the number of biases of layer $i,~i\in\{1,\dots,d\}$. The network parameters are defined according to
\begin{eqnarray*}
\tau_{wi} &:=& log(1+exp(\delta_{wi}))\\
\mathcal{W}_{i} &:=& \underline{\mu}_{wi}\odot(\underline{1}_{k_{i}}+\tau_{wi}\underline{\varepsilon}_{wi})\\
\tau_{bi} &:=& log(1+exp(\delta_{bi}))\\
\mathcal{B}_{i} &:=& \underline{\mu}_{bi}\odot(\underline{1}_{k^{\prime}_{i}}+\tau_{bi}\underline{\varepsilon}_{bi}),
\end{eqnarray*}
where $\underline{1}_{k}$ is the $k$-dimensional vector consisting of all ones and $\odot$ is the Hadamard or elementwise product. The variational parameters are represented by $\underline{\mu}_{wi}\in\mathbb{R}^{k_{i}}$, $\underline{\mu}_{bi}\in\mathbb{R}^{k^{\prime}_{i}}$, $i\in\{1,\dots,d\}$ and $\delta_{wi},~\delta_{bi}\in\mathbb{R}$. We choose $\varepsilon_{wi}\in\mathbb{R}^{k_{i}}$ and $\varepsilon_{bi}\in\mathbb{R}^{k^{\prime}_{i}}$ to be multivariate standard normally distributed. Thus, the network weights and biases are also normally distributed,
\begin{eqnarray}
\mathcal{W}_{i} &\sim & \mathcal{N}(\underline{\mu}_{wi},~\tau_{wi}^{2}~\cdot diag(\underline{\mu}_{wi}))\\
\mathcal{B}_{i} &\sim & \mathcal{N}(\underline{\mu}_{bi},~\tau_{bi}^{2}~\cdot diag(\underline{\mu}_{bi})).
\end{eqnarray}
The diagonal covariance matrix of the Gaussian variational distributions assumes the independence of all the network parameters and biases. The calculation of the respective gradients and an extension to a tridiagonal covariance matrix can be found in~\citep{steinbrener2020measuring} and~\citep{posch2020correlated}. We implement the Bayesian segmentation network using the open source Python library PyTorch~\citep{paszke2017automatic}.

\subsection{Pose Estimation}
\label{subsec:workflow-pose-estimation}
In order to generate a static simulation model of an assembly plant, the pose of each object of interest within the segmented point cloud needs to be determined. Pose estimation is carried out for each class $j,~j\in\{1,\dots,m\}$ separately. The number of objects within each class is determined and the points belonging to the individual objects are identified using a clustering approach. Technically, any clustering algorithm can be applied, however, based on the data set characteristics it makes sense to determine what algorithm suits our use case best. Thus, we evaluate the classical clustering methods $k$-means~\citep{macqueen1967some} and fuzzy $c$-means~\citep{bezdek1984fcm}, the density based approaches "Density-Based Spatial Clustering of Applications with Noise"~(DBSCAN)~\citep{ester1996density} and "Ordering Points To Identify the Clustering Structure"~(OPTICS)~\citep{ankerst1999optics} as well as spectral clustering~\citep{ng2002spectral} in Section~\ref{subsubsec:eval-sim-gen}.\\
After knowing the number of objects within each class their pose needs to be determined. A prerequisite to tackle this problem is the availability of 3D CAD models or at least reference point clouds of the object classes of interest. A point cloud can be sampled from these CAD models resulting in a "perfect" point cloud of an object within class~$j$. Iteratively an initial object pose is determined for every object within each class by coarsely registering the "perfect" or reference point cloud and one object of the segmented point cloud. This can be achieved by the global registration algorithm "Random Sample \mbox{Consensus"~(RANSAC)~\citep{fischler1981random}}, which does not require an initial alignment. A tight alignment of the point clouds is achieved using the local registration algorithm ICP that interprets the output of the RANSAC algorithm as an initial alignment. Based on the resulting transformation matrix the respective (x,y,z)-coordinates and the roll, pitch and yaw are calculated for each of the objects. These values are retained for later placement within a simulation environment. The object details are expressed in the Automation Markup Language~(AML) format, which is an open standard to exchange factory engineering data. Thus, it seamlessly fits into a holistic data model. A pseudo code of pose estimation is provided in Algorithm~\ref{alg:pose-estimation}. In the pseudo code we assume that the clustering algorithm is capable of determining the number of cluster centres itself. Otherwise, the number of clusters has to be passed to the algorithm as input parameter. Further, we assume the availability of CAD models out of which reference point clouds are generated.  

\begin{algorithm}
\SetAlgoLined
\textbf{Input:} \textit{m}... number of classes, \textit{pc\_seg}... segmented point cloud,\\
~~~~~~~~~~~~\textit{CAD\_models}... CAD geometry belonging to each class of interest\\
\textbf{Output:} \textit{object\_poses.aml}... an aml-file containing each object pose\\
~\\
\ForEach{class $j\in\{1,\dots, m\}$}
{
$c$~~ $\gets$ number of cluster centres of class $j$ given by the clustering algorithm\\
$pc$ $\gets$ distinct point clouds of $c$ objects in class \textit{j} derived by a clustering\\
~~~~~~~~~~ routine of class $j$ within $pc\_seg$\\
$pc\_ perf$ $\gets$ sampled reference point cloud from CAD model of class $j$\\
~\\

\ForEach{object $pc[l]$ of class $j$, $l\in \{1,\dots, c\}$}
{
$transf\_init$ $\gets$ transformation~matrix~of~RANSAC~registration of $pc[l]$\\
~~~~~~~~~~~~~~~~~~~~~~~~~~and $pc\_perf$\\
$transf\_final$ $\gets$ updated transformation matrix of ICP registration of\\
~~~~~~~~~~~~~~~~~~~~~~~~~~~~~$pc[l]$ and $pc\_perf$ initialized with the RANSAC\\ ~~~~~~~~~~~~~~~~~~~~~~~~~~~~~alignment $transf\_init$\\
$object\_pose$ $\gets$ calculate object pose based on $transf\_final$\\
$object\_poses.aml$ $\gets$ append $pc[l]$ including $object\_pose$
}
}
\KwRet{$object\_poses.aml$}
\caption{Pose estimation($m,~pc\_seg,~CAD\_ models$)}
\label{alg:pose-estimation}
\end{algorithm}

\subsection{Simulation Model Generation}
The extracted object poses can be used to generate a simulation model in any simulation environment. For displaying our final factory model, we choose the simulation framework Unreal~Engine~4~(UE4)~\citep{Unreal} as this simulation tool is widely applied in different industries due to its Blueprints visual scripting system, which is easy to use even for people without software engineering background. Though it is also possible to develop simulations using common programming and scripting languages like C++ or Python. Prerequisite for the simulation model generation is the availability of the CAD models of every class of interest and an Unreal project having access to these CAD geometries. Of course, also point clouds of objects, which are not available in CAD, can be included in the simulation model. However, as point clouds are not complete, i.e. they have holes due to occlusions during digitalization, the informative value of the simulation can suffer. For instance, collision checking is not reliable in the presence of potentially incomplete point clouds. Using the positional and rotational data saved in the \mbox{\textit{object\_ poses.aml}} file the CAD object is placed in the scene. The placement itself in the UE4 environment is straight forward.

\section{Results and Analysis}
\label{sec:evaluation-and-analysis}
This section discusses the properties of the collected automotive factory data set, which is generated and pre-processed according to the workflow described in \mbox{Section~\ref{sec:workflow-description}} and used for evaluation of the data processing steps. The segmentation network is further evaluated on the publicly available Stanford Large-Scale 3D Indoor Spaces data set for comparability reasons. In the sequel the results and recommended actions for our proposed workflow are presented. The Bayesian neural network itself is evaluated in our previous work~\citep{petschnigg2020uncertainty}. Thus, we will briefly summarize the main results and evaluate the remaining process steps in more detail.

\subsection{Evaluation Data Sets}
The complete workflow is evaluated on the data set we collect according to the presented methodology. This includes the evaluation of data collection, segmentation and pose estimation. In order to ensure the comparability of the segmentation results of the BNN to other state-of-the-art segmentation frameworks the publicly available Stanford Large-Scale 3D Indoor Spaces Data Set is used.

\subsubsection{Stanford Large-Scale 3D Indoor Spaces Data Set}
The Stanford large-scale 3D indoor spaces (S3DIS) data set~\citep{armeni20163d} is a point cloud data set of 6 indoor spaces, which are spread across three different buildings. It represents an area of over \mbox{6 000 m$^{2}$} as a point cloud with more than 215 million points. It comprises 13 distinct classes of structural and furniture elements that belong to educational facilities, offices, hallways and sanitary installations. This data set is labelled on instance level and can be downloaded from \url{http://buildingparser.stanford.edu/dataset.html}.

\subsubsection{Automotive Factory Data Set}
\label{subsubsec:factory-data-set}
\begin{figure}[h]
\centering
\begin{subfigure}[c]{0.48\textwidth}
\includegraphics[width=0.95\textwidth]{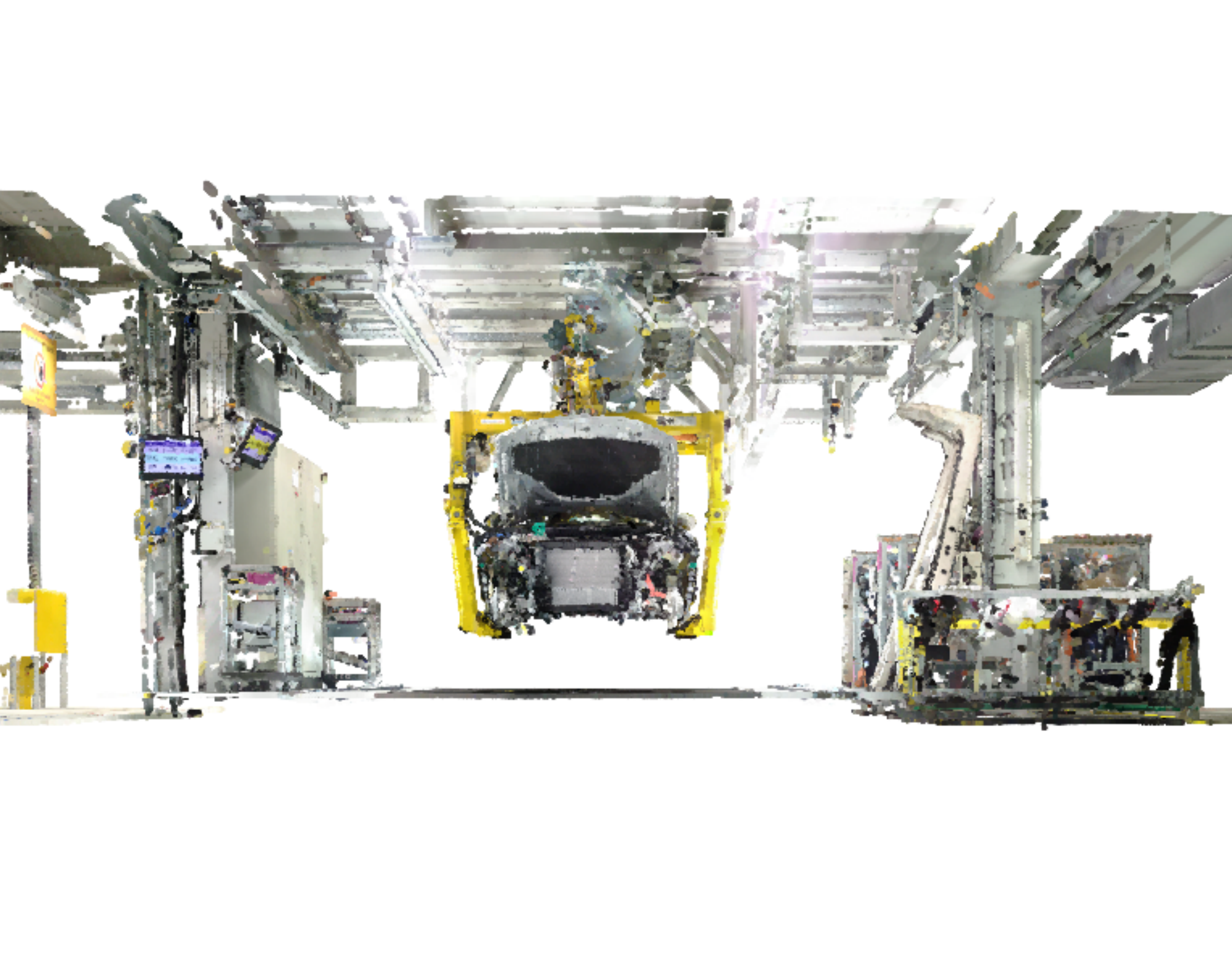}
\subcaption{}
\end{subfigure}
\begin{subfigure}[c]{0.48\textwidth}
\includegraphics[width=0.95\textwidth]{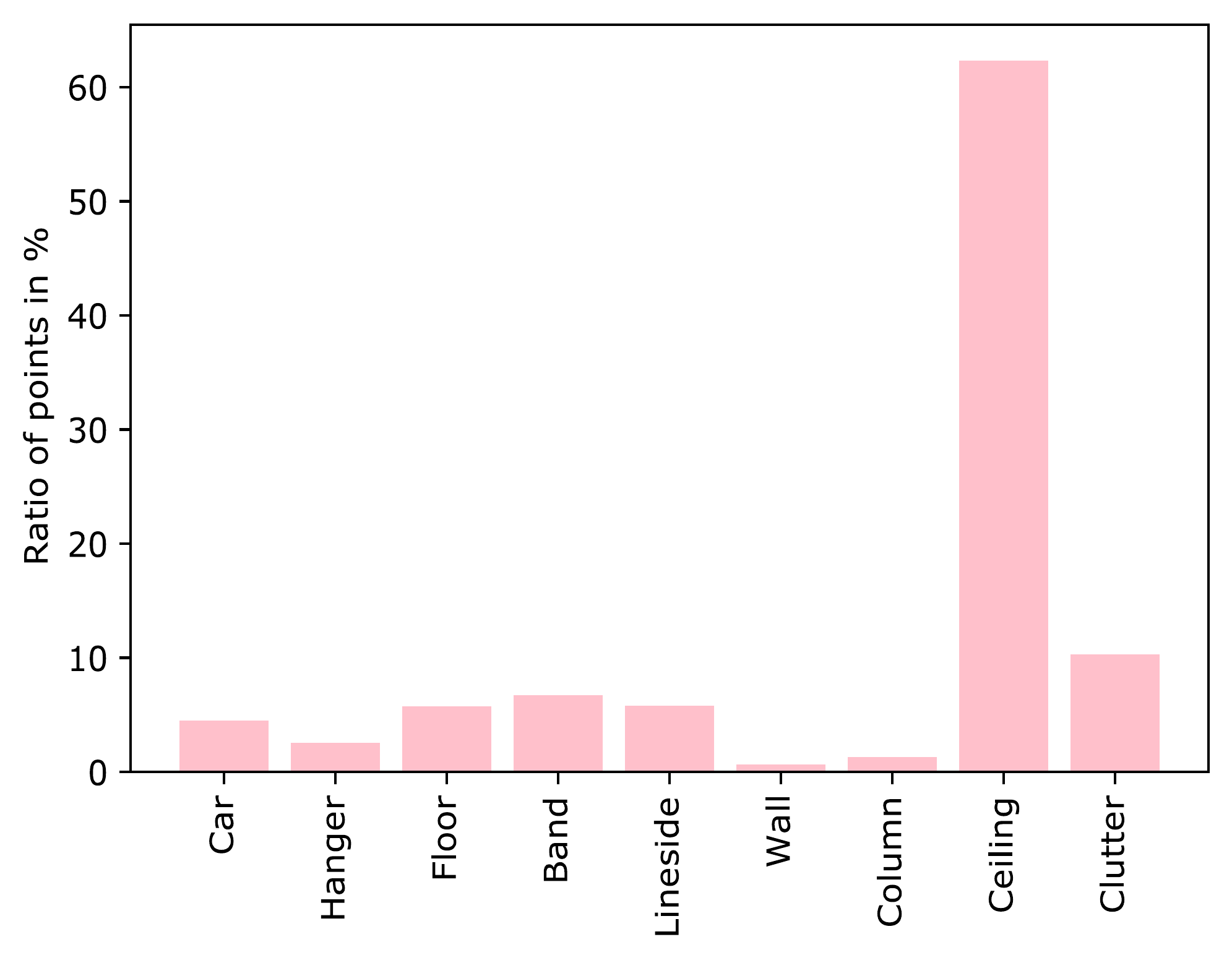}
\subcaption{}
\end{subfigure}
\caption{Summary of the generated point cloud. (\textbf{a}) One example tact of the collected data set. (\textbf{b}) Highly imbalanced class distribution.}
\label{fig:automotive-data-set}
\end{figure}

The automotive factory data set is collected according to the methods described in Section~\ref{subsec:data-collection}. On the basis of several static laser scans as well as thousands of camera images we generate local laser scan point clouds as well as photogrammetric point clouds for 13 tacts of car body assembly. These local point clouds are registered using registration targets to form one global point cloud, which includes over one billion points initially. The point cloud is cleaned using noise reducing filters. The remaining noise points are removed during the labelling steps using the open source software tool CloudCompare~\citep{CloudCompare}. The final point cloud comprises \mbox{594,147,442 points}, which corresponds to a removal of about 40\% of the points compared to the raw point cloud. The majority of the noise points are the result of motion blur that occurs during scanning and reflections of shinily painted objects. In total the point cloud is divided into 9 classes comprising 8 object categories relevant for factory planning as well as the additional class of clutter, which serves as a placeholder for all the other objects in the scene. The 8 object classes correspond to car, hanger, floor, band, lineside, wall, column and ceiling. These are relevant for factory planning as they are difficult to move or even immovable. One example tact of the collected data set is illustrated in \mbox{Figure~\ref{fig:automotive-data-set}~(a)}. After labelling all the data, the class distribution can be analysed. We see that the classes are highly imbalanced, which is depicted in \mbox{Figure~\ref{fig:automotive-data-set}~(b)}.

\begin{figure}[h]
\centering
\begin{subfigure}[b]{0.85\textwidth}
   \includegraphics[width=1\linewidth]{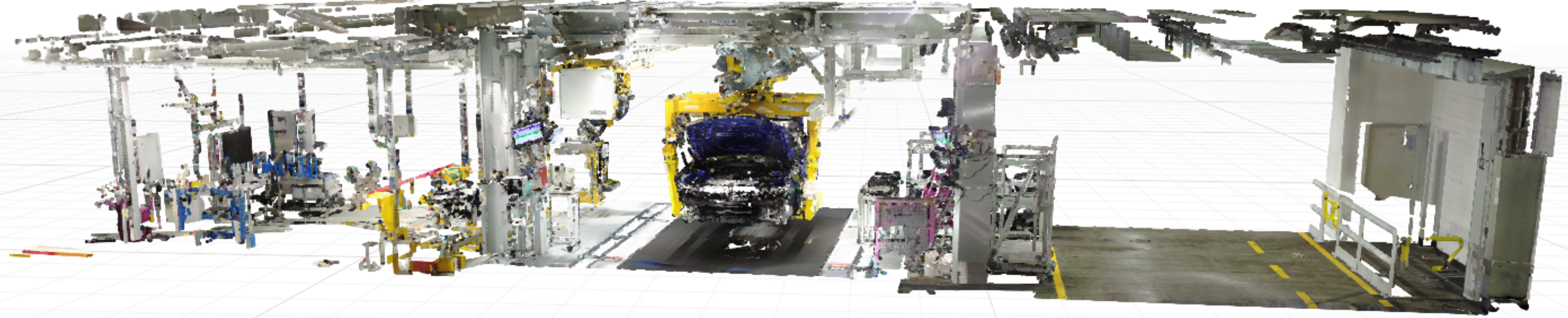}
   \caption{}
\end{subfigure}
\vspace{0.65\floatsep}
\vspace{0.1\floatsep}
\begin{subfigure}[b]{0.85\textwidth}
   \includegraphics[width=1\linewidth]{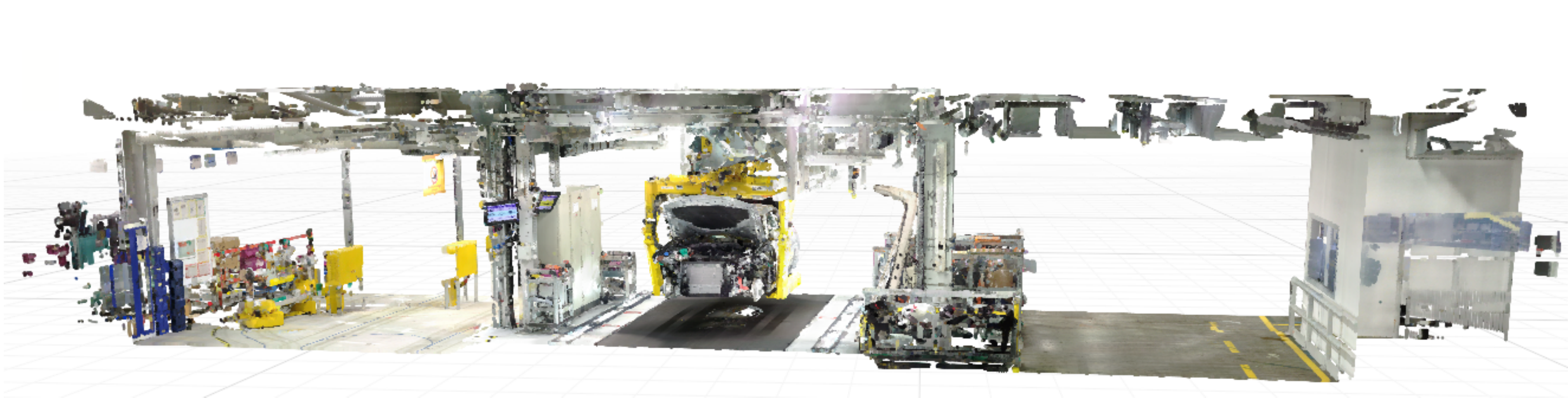}
   \caption{}
\end{subfigure}
\caption{The two test tacts of the automotive factory data set.}
\label{fig:test-tacts}
\end{figure}

It is clear that a large ratio of points is assigned to the class ceiling, whereas the classes of wall and column comprise relatively few points. The size of the class ceiling stems from the layered architecture of this structure. In response to this, points on various levels belong to this class. In contrast to that walls and columns are hung with clutter objects like information signs, fire extinguishers or rubbish bins. Therefore, only few points truly belong to these classes and their point clouds have many holes due to occlusions caused by the clutter objects.\\
Out of the 13 digitalized assembly tacts two are set aside for evaluating different steps of the workflow. The remaining tacts are used for model training and validation. Figure~\ref{fig:test-tacts} displays the two test tacts.

\subsection{Results}
\label{sec:results}
The complete workflow of simulation model generation is evaluated in the context of automotive car body assembly. The analysis starts with evaluating different digitalization approaches in terms of point cloud accuracy, completeness and point density. The evaluation of the Bayesian segmentation network is only summarized as a more detailed analysis can be found in~\citep{petschnigg2020uncertainty}. Additionally, different clustering algorithms are analysed for suitability in our workflow. This analysis takes into account the clustering accuracy and time. The accuracy of the pose estimation step is evaluated with respect to the mean deviation of the generated simulation scene to the reference point cloud.

\subsubsection{Data Collection and Visualization}
\label{subsec:results-data-coll}
For the evaluation of our workflow, we first collect the automotive factory data set and analyse the goodness of the resulting point cloud with respect to its accuracy, completeness and point density. In order to calculate the global accuracy of the measured point cloud we suppose that the distance of the measured points to the reference model is normally distributed with mean zero. The variance is given by

\begin{equation}
\sigma^{2}~=~\frac{1}{n}\sum_{i=1}^{n}(d_{i}-0)^{2}~=~\frac{1}{n}\sum_{i=1}^{n}d_{i}^{2},
\end{equation}

where $d_{i}$ corresponds to the measured distance of point $i$ of the measured point cloud to the reference model. The standard deviation $\sigma$ is the square root of the variance and is referred to as accuracy of the point cloud, i.e. a lower standard deviation corresponds to a higher accuracy. In order to calculate point cloud completeness, the minimum distance of each point in the measured point cloud to the reference model is calculated. Then we define completeness as the percentage of points with a maximum distance of 10~mm to the reference point cloud. The point density is quantified by the mean number of points within a 10~mm radius of each point in the measured point cloud.\\
We generate point clouds from varying sources, i.e. four laser scans per tact, six laser scans per tact, photogrammetry only using a fish-eye and a wide-angle lens, a combination of four laser scans with photogrammetry as well as six laser scans with photogrammetry. The evaluation results for the point clouds generated from these sources are summarized in Table~\ref{tab:eval-visu}. The unit of point density is the mean number of points.

\begin{table}[h] 
\centering
\caption{Evaluation of the collected point cloud in terms of accuracy, completeness and point density.}
\begin{tabular}{lccc}
\hline
\textbf{Data Source} & \textbf{Accuracy} & \textbf{Completeness} & \textbf{Point Density}\\
\hline
4 Scans & $\pm$5.3 mm & 41\% & 89.4 \\
6 Scans & $\pm$4.8 mm & 51\% & 146.7 \\
Photogrammetry only wide-angle & $\pm$5.8 mm & 61\% & 854\\
Photogrammetry only fish-eye & $\pm$7.8 mm & 60\% & 154.2\\
4 Scans and Photogrammetry & $\pm$6.3 mm & 61\% & 804.3\\
6 Scans and Photogrammetry & $\pm$6.2 mm & 64\% & 789.9\\
\hline
\end{tabular}
\label{tab:eval-visu}
\end{table}

\begin{figure}[h]
\centering
\begin{subfigure}{0.30\textwidth}
  \includegraphics[width=\linewidth]{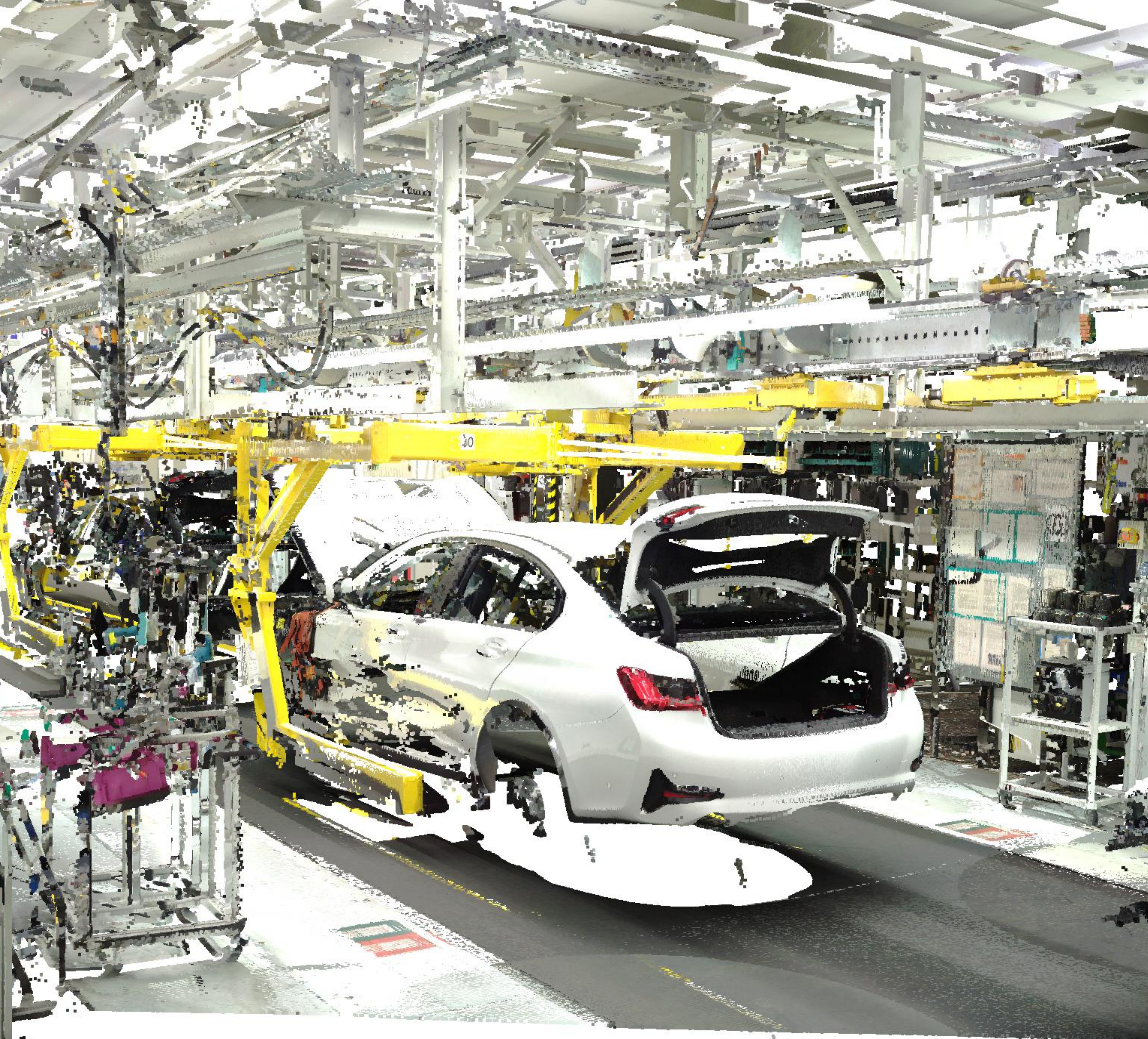}
  \caption{}
\end{subfigure}\hfil 
\begin{subfigure}{0.30\textwidth}
  \includegraphics[width=\linewidth]{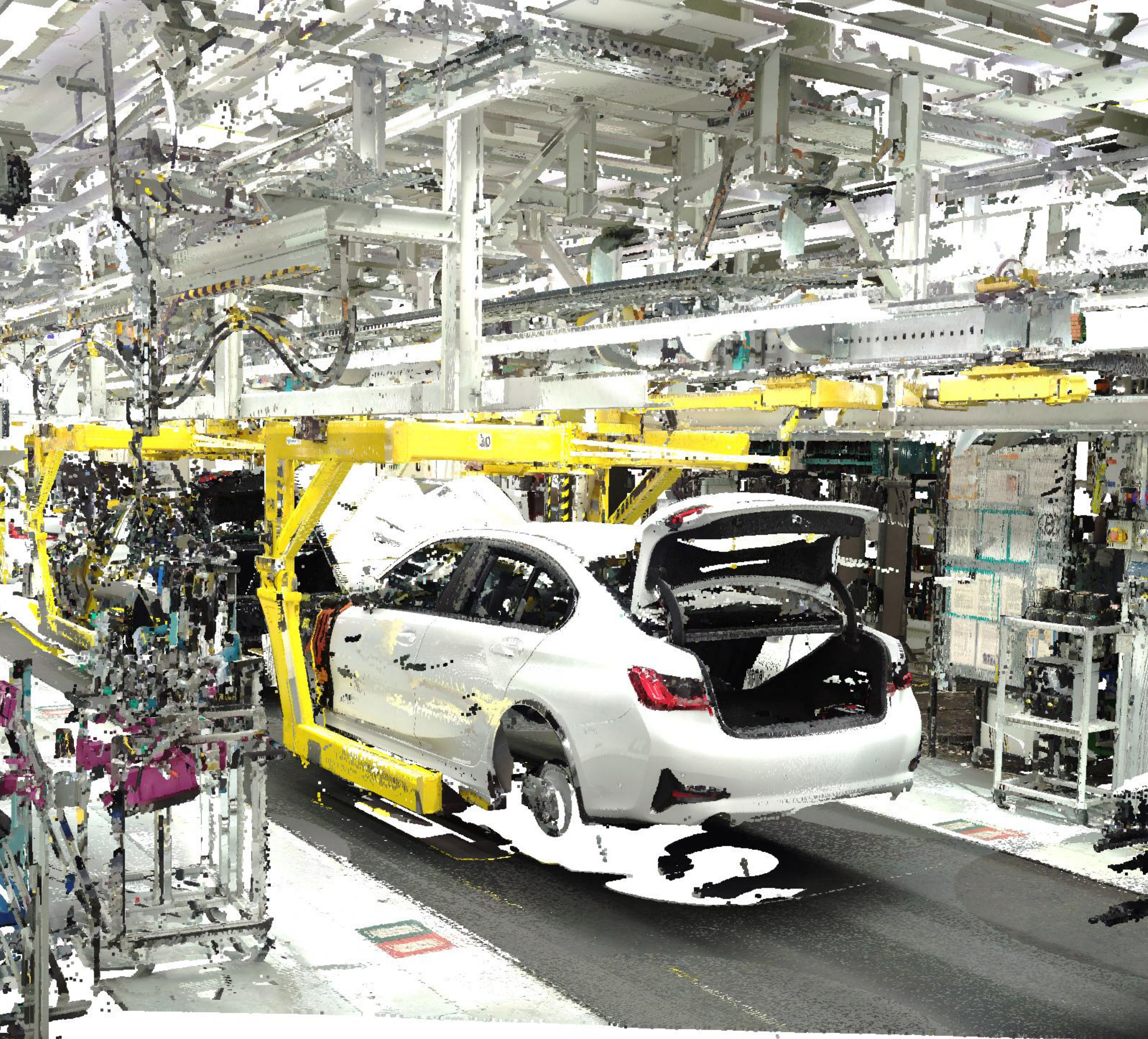}
  \caption{}
\end{subfigure}\hfil 
\begin{subfigure}{0.30\textwidth}
  \includegraphics[width=\linewidth]{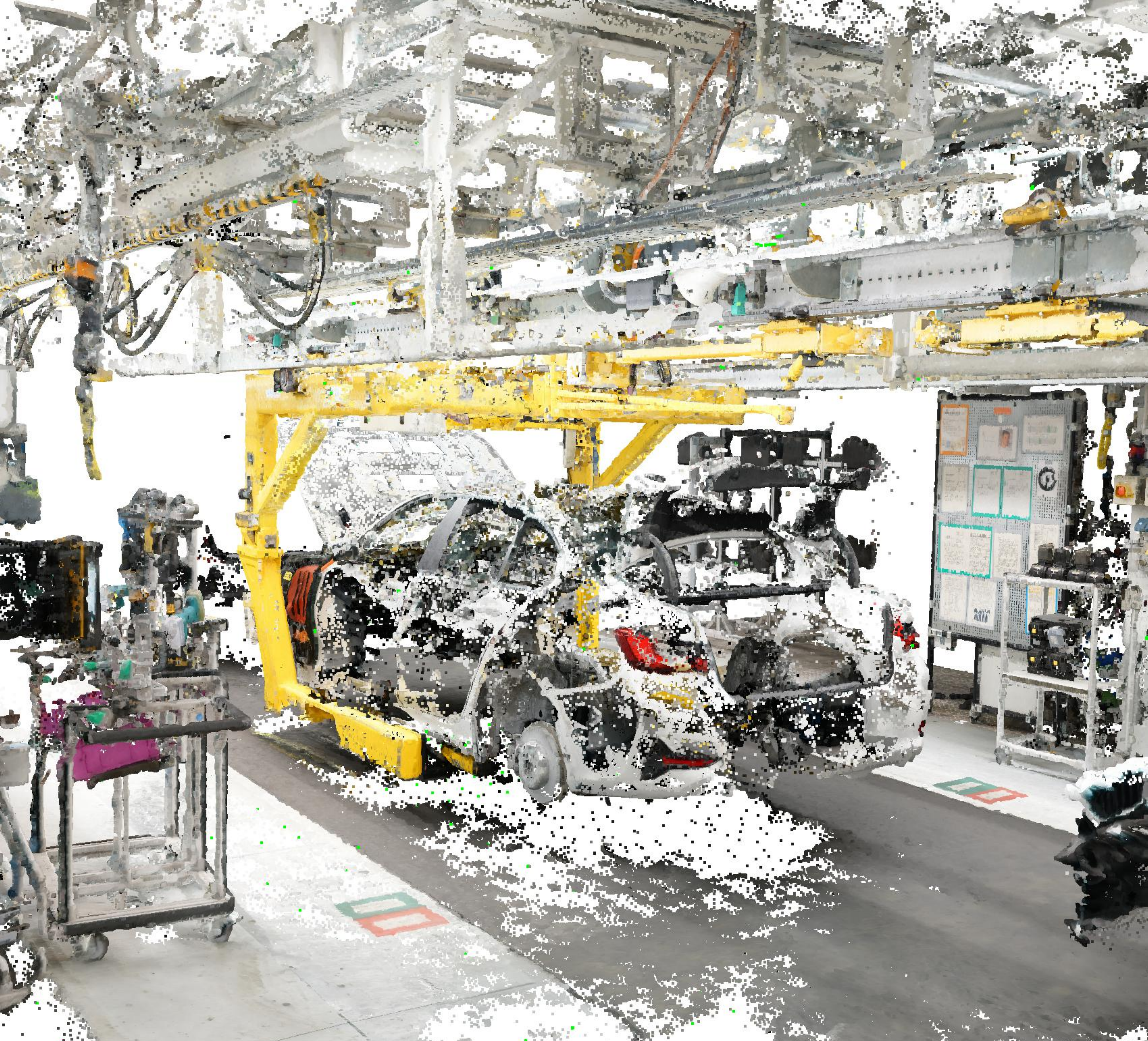}
  \caption{}
\end{subfigure}

\medskip
\begin{subfigure}{0.30\textwidth}
  \includegraphics[width=\linewidth]{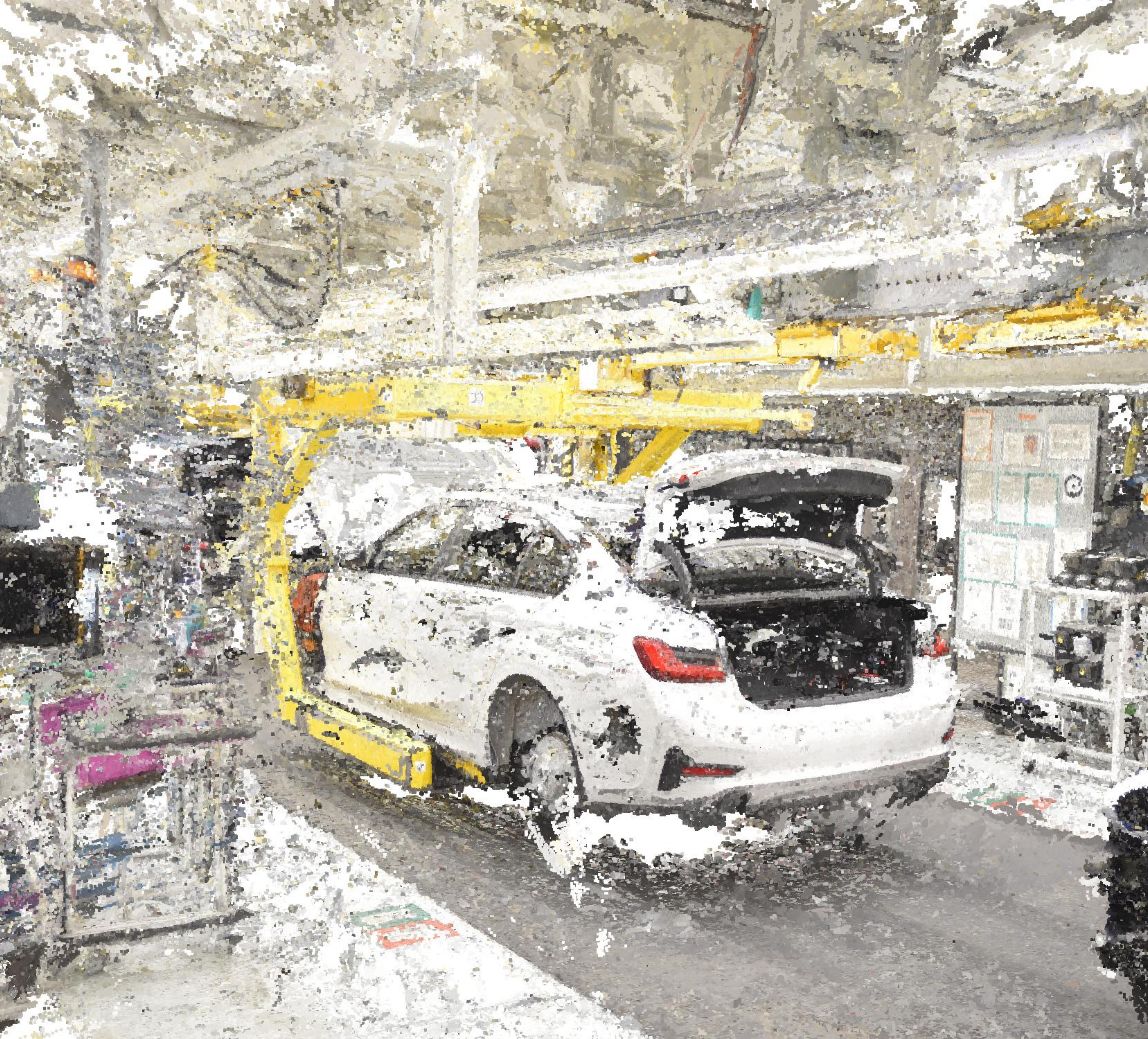}
  \caption{}
\end{subfigure}\hfil 
\begin{subfigure}{0.30\textwidth}
  \includegraphics[width=\linewidth]{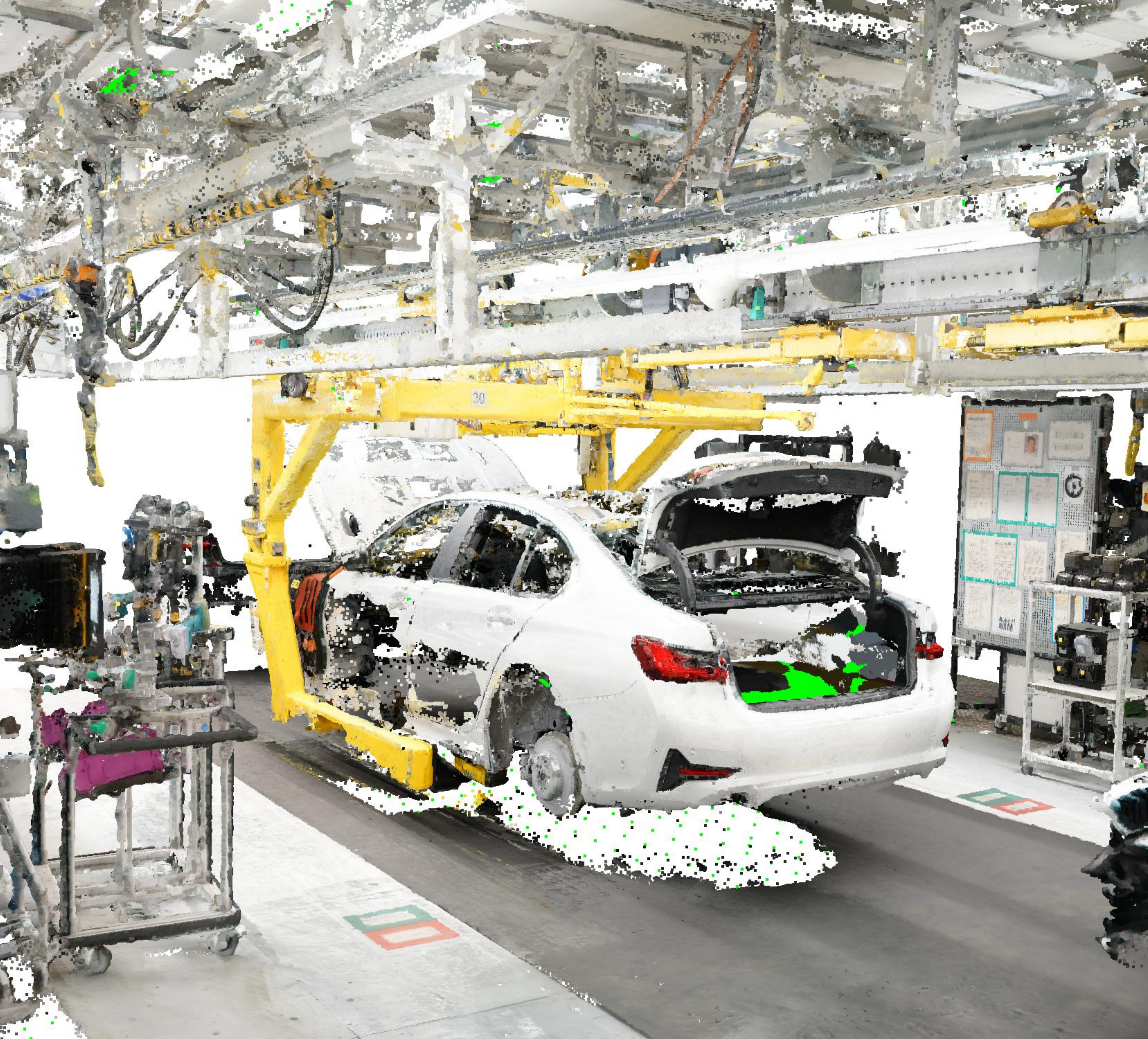}
  \caption{}
\end{subfigure}\hfil 
\begin{subfigure}{0.30\textwidth}
  \includegraphics[width=\linewidth]{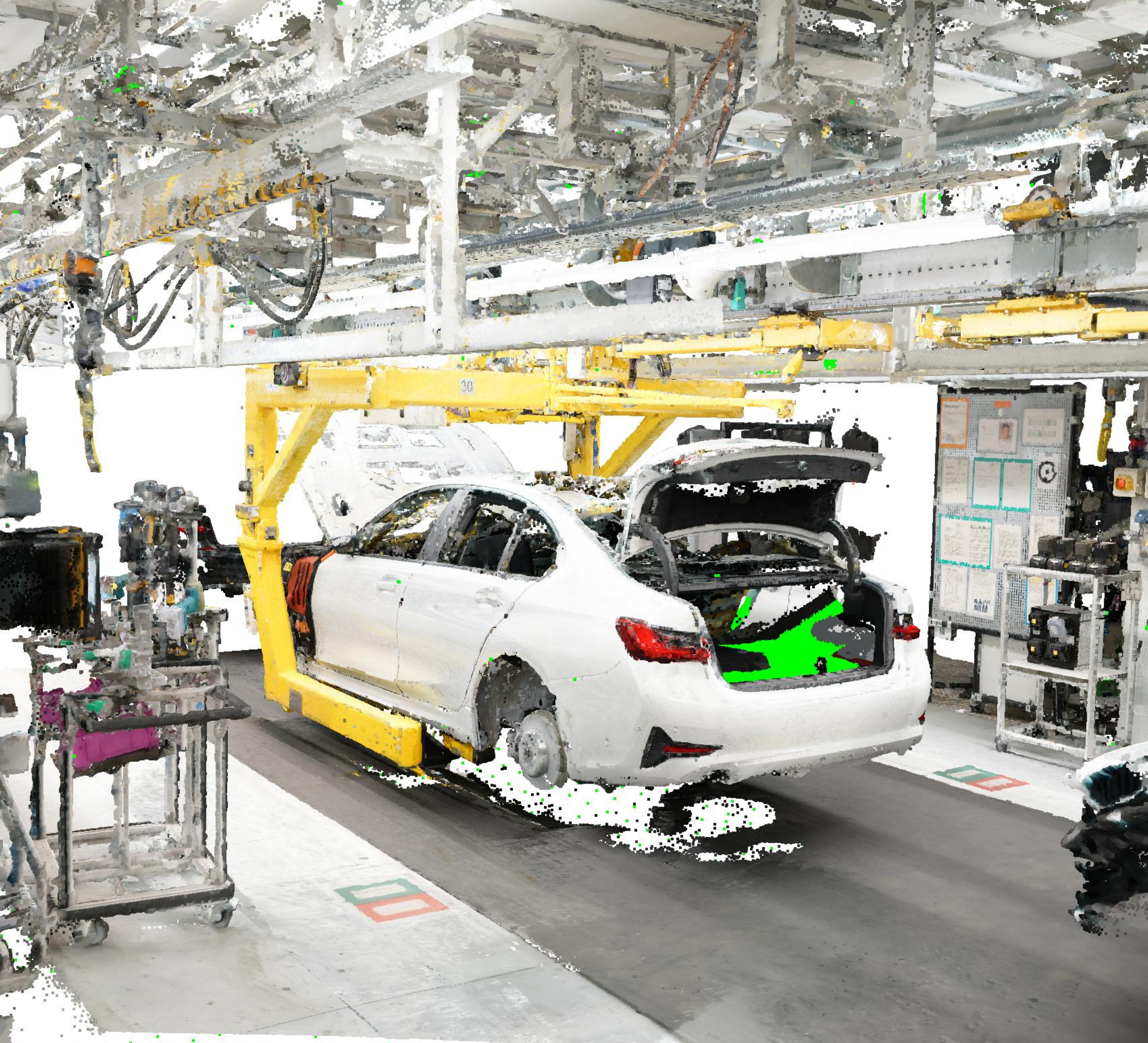}
  \caption{}
\end{subfigure}
\caption{Point clouds resulting from different data sources. \textbf{(a)}~Four laser scans. \textbf{(b)}~Six laser scans. \textbf{(c)}~Photogrammetry with wide-angle lens. \textbf{(d)}~Photogrammetry with fish-eye lens. \textbf{(e)}~Four laser scans and photogrammetry. \textbf{(f)}~Six laser scans and photogrammetry.}
\label{fig:evaluation-visu}
\end{figure}

We see that the laser scan only point clouds have a higher accuracy, which corresponds to a lower standard deviation, than point clouds including photogrammetry information. In contrast to that, completeness and point density increase considerably when the photogrammetric information is added to the laser scans. Surprisingly, the point density is highest for the photogrammetry only case using the wide-angle lens. Further, the point density is higher in the case of four laser scans and photogrammetry compared to six laser scans and photogrammetry. Both of the effects are dependent on the number of blurred images that are removed and the quality of image registration. The more images are used for generating the photogrammetric point cloud the higher the resulting point density.\\
For the generation of a simulation model the point density is neglectable as the point clouds are down-sampled before segmentation for efficiency reasons. However, a high accuracy is important in order to ensure correct model placement. Thus, there are different recommended processes for data collection depending on the intended use of the point cloud. \mbox{Figure~\ref{fig:evaluation-visu}} shows all the point clouds generated from the given data sources. In \mbox{Figure~\ref{fig:evaluation-visu}~(a)} and~(b) only the static laser scanner is used taking four and six scans per tact, respectively. The two pure photogrammetric point clouds are illustrated in \mbox{Figure~\ref{fig:evaluation-visu}~(c)} and~(d), which we generate using the wide-angle and the fish-eye lens. The point clouds generated by four and six laser scans complemented with the information of the photogrammetric point cloud are shown in \mbox{Figure~\ref{fig:evaluation-visu}~(e)} and~(f). For the combined laser scan and photogrammetric point clouds the wide-angle lens is used.

\begin{figure}[h]
\centering
\includegraphics[width=0.85\textwidth]{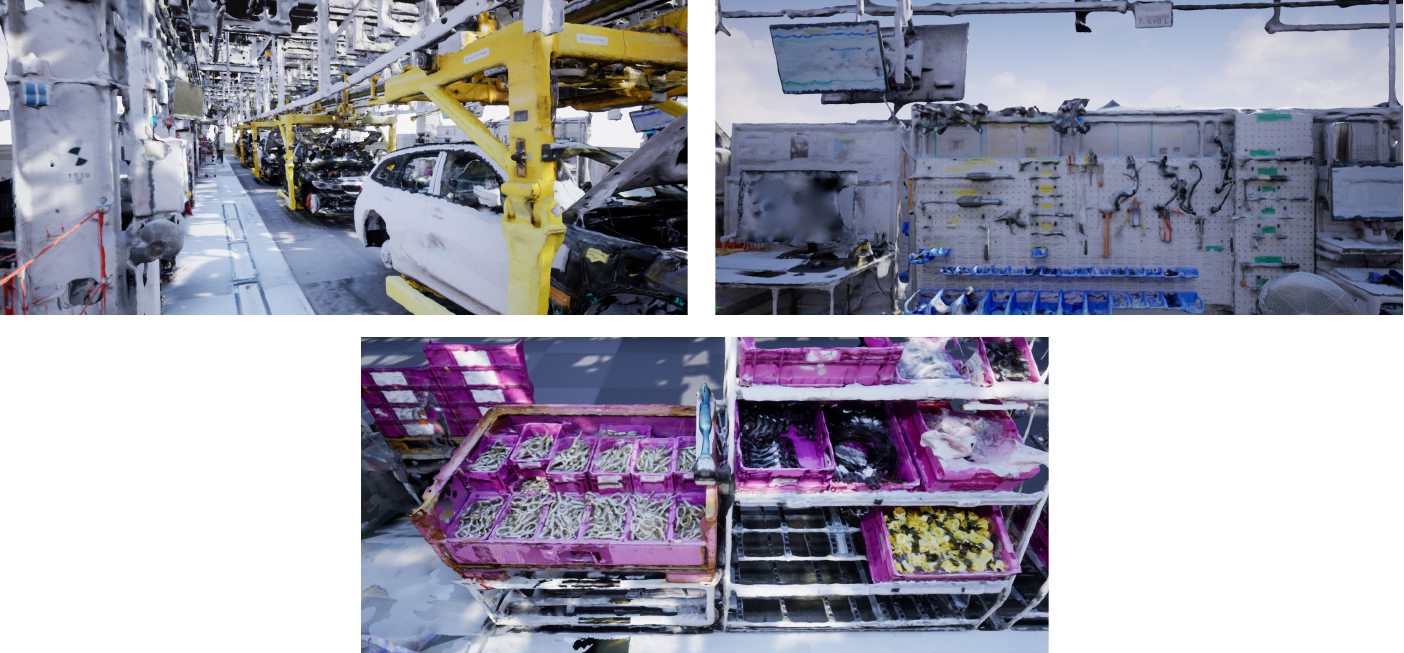}
\caption{Visualization of the combined laser scan and photogrammetric point cloud as meshes.}
\label{fig:evaluation-visu-mesh}
\end{figure}

From a visual inspection, we can see that the point cloud using six laser scans is better in terms of completeness, especially at the side of the car where reflections cause holes in the point cloud generated from four scans. The photogrammetric point clouds suffer from reflections due to the shiny car paint as well. Further, the photogrammetric point cloud generated with the fish-eye lens is distorted, which can also be seen in the higher standard deviation. Generally, images that are taken using a fish-eye lens are easier to register due to a higher number of features being captured on the image. However, cameras with a fish-eye lens are more difficult to calibrate due to the high degree of distortion and the resulting point cloud has a lower accuracy compared to the wide-angle lens. The combined laser scan and photogrammetric point clouds look best in terms of completeness whereby again the point cloud produced with six laser scans has a higher quality.\\
The software tool RealityCapture~\citep{RealityCapture} can be used for quickly visualizing the generated point cloud data as a triangle mesh. In order to improve the handling of the mesh the number of triangles is reduced in Blender~\citep{Blender}. The visualization of the optimized meshes is depicted in Figure~\ref{fig:evaluation-visu-mesh}. The displayed meshes are blocks in which objects are not separated from each other. Thus, they are not directly usable for factory or process simulation since single objects can neither be handled nor animated. However, these visualizations serve as a good starting point for factory planners that are not on-site to get an overview of the current plant layout. Note that we do not aim at creating photo realistic meshes out of point clouds in this work, which involves high efforts in modelling and texturizing the scene.

\subsubsection{Bayesian Segmentation}
\label{subsubsec:result-bayesian-segmentation}
After generating a global point cloud of the factory environment, the segmentation step can be evaluated. To this end, the Bayesian formulation of PointNet is compared to the original PointNet implementation of~\citep{qi2017pointnet} based on TensorFlow~\citep{abadi2016tensorflow} and our PyTorch~\citep{paszke2017automatic} implementation of frequentist PointNet. As already mentioned, the automotive factory data set that is collected according to the presented workflow is used for network evaluation. In order to ensure comparability to other segmentation frameworks the publicly available S3DIS data set is used. The input point clouds comprising rooms or assembly tacts are cut into equally sized blocks, which serve as an input for all the networks. The number of points within these blocks is sampled to 4096.\\
The models are trained using mini-batch stochastic gradient descent with a batch size of 16 on both data sets. The momentum parameter is set to 0.9 and a decaying learning rate~$lr$ is used with an initial learning rate of $lr=0.001$ in the frequentist and $lr=0.01$ in the Bayesian case. The learning rate is decayed every 10 epochs by a factor of $0.7$ during frequentist training and $0.9$ during Bayesian training. The batch size and the learning rate are optimized by using grid search and cross-validation. In terms of approximation of the posterior predictive distribution, we draw $K=50$ Monte Carlo samples.\\
Both models converge and the training is stopped after 100~epochs. Table~\ref{tab:class-vs-bayes} displays the evaluation results of the original PointNet, the frequentist PyTorch implementation and our Bayesian PointNet on area~5 of the S3DIS data set. Further, the networks are evaluated on the test tacts of our automotive factory data set. Bold table entries represent best model performance.

\begin{table}[h] 
\caption{Segmentation results of the original, classical and Bayesian PointNet on S3DIS and our automotive factory data set. For every model the training converges and is stopped after 100 epochs.}
\centering
\begin{tabular}{ccccc}
\hline
\textbf{Model} & \textbf{Data} & \textbf{Test Acc.} &  \textbf{Test mIoU}\\
\hline
Original PointNet \citep{qi2017pointnet} & S3DIS & 78.62\% & 0.4771 \\
Classical PointNet & S3DIS  & 79.81\% & 0.4374 \\
Bayesian PointNet & S3DIS  & \textbf{81.16\%} & \textbf{0.4834} \\
\hline
Classical PointNet & Automotive & 94.23\% & 0.7808\\
Bayesian PointNet & Automotive & \textbf{95.47\%} & \textbf{0.8263}\\
\hline
\end{tabular}
\label{tab:class-vs-bayes}
\end{table}

It can be seen that Bayesian PointNet clearly outperforms the original implementation as well as the classical version in terms of segmentation accuracy and the mean intersection over union (IoU) for both data sets. For a more detailed review and interpretation of the segmentation results see~\citep{petschnigg2020uncertainty}.

\subsubsection{Uncertainty Estimation}
Similarly to the evaluation of the segmentation results, we only provide a brief summary of the evaluation of uncertainty estimation, which is discussed in more detail in~\citep{petschnigg2020uncertainty}. The different methods used for uncertainty estimation are described in Section~\ref{subsec:BNN-Unc}. They include entropy based estimation methods, the predictive variance as well as the estimation of credible intervals on the network outputs. Each of these quantities are calculated on the basis of $K=50$ forward passes through the network each using a different weight sample drawn from the variational distribution. Predictive and aleatoric uncertainty are estimated using the Equations~\ref{eq:U-pred} and~\ref{eq:U-alea}. Epistemic uncertainty is the difference of these two values. In order to determine the predictive variance, the unbiased estimator of the variance is used. In the credible interval based method we determine the 95\%-credible intervals of the predictive network outputs after normalization with the softmax function.\\
In order to determine how uncertainty improves network performance, we compare the accuracy of the Bayesian baseline to the accuracy achieved when only considering certain points with respect to all the discussed uncertainty measures, i.e. uncertain points are dropped. This time area 6 of the S3DIS data set is used for evaluation as it is smaller and thus more efficient to test on. The results for one of each room types in area 6 of the S3DIS data set and both test tacts of the automotive factory data set are summarized in Table~\ref{tab:eval-unc-rooms-tacts}.

\begin{table}[h]
\caption{Evaluation of the discussed methods for uncertainty estimation and influence on model accuracy.}
\centering
\begin{tabular}{ccccccc}
\hline
 & \textbf{Baseline} & \textbf{Predictive} & \textbf{Aleatoric} & \textbf{Epistemic} & \textbf{Variance} & \textbf{Credible Int.}\\
\hline
\textbf{Conf. Room} & 88.92\% & 92.56\% & \textbf{92.59\%} & 91.03\% & 90.82\% & 91.01\%\\
\textbf{Copy Room} & 70.82\% & 72.56\% & 72.58\% & 72.37\% & 72.02\% & \textbf{74.85\%}\\
\textbf{Hallway} & 81.13\% & 83.01\% & 82.98\% & 82.98\% & 83.20\% & \textbf{93.06\%}\\
\textbf{Lounge} & 71.77\% & 73.31\% & 73.49\% & 73.09\% & 73.06\% & \textbf{77.16\%}\\
\textbf{Office} & 90.77\% & \textbf{93.02\%} & 93.01\% & 92.28\% & 92.37\% & 92.42\%\\
\textbf{Open Space} & 76.68\% & 79.45\% & 79.54\% & 77.96\% & 77.77\% & \textbf{80.38\%}\\
\textbf{Pantry} & 76.21\% & 78.51\% & 78.58\% & 77.44\% & 77.24\% & \textbf{79.20\%}\\
\hline
\textbf{Assembly T. 1} & 94.21\% & 96.63\% & \textbf{96.64\%} & 95.54\% & 95.60\% & 94.99\%\\
\textbf{Assembly T. 2} & 94.78\% & 97.41\% & \textbf{97.47\%} & 96.22\% & 96.22\% & 95.63\%\\
\hline
\end{tabular}
\label{tab:eval-unc-rooms-tacts}
\end{table}

For the entropy based measures as well as the predictive variance a threshold needs to be set, in order to classify a prediction as either certain or uncertain. In this case we consider predictions as certain when the respective uncertainty value is smaller or equal to the mean uncertainty plus two sigma. For the credible interval based method, predictions are considered certain if the 95\%-credible interval of the predicted class does not overlap with the 95\%-credible interval of any other class. It is clear that considering any of the uncertainty estimation approaches considerably increases the segmentation accuracy. In Table~\ref{tab:uncertainty-points-bayes} the ratio of points that are considered uncertain by the respective methods are displayed.

\begin{table}[h]
\caption{Percentage of predictions dropped when excluding uncertain predictions.}
\centering
\begin{tabular}{ccccccc}
\hline
& \textbf{Baseline} & \textbf{Predictive} & \textbf{Aleatoric} & \textbf{Epistemic} & \textbf{Variance} & \textbf{Credible}\\
\hline
\textbf{Conf. Room} & - & 6.86\% & 6.90\% & 5.49\% & 5.50\% & 4.07\% \\
\textbf{Copy Room} & - & 3.25\% & 3.28\% & 4.50\% & 4.81\% & 9.53\% \\
\textbf{Hallway} & - & 4.64\% & 4.65\% & 5.08\% & 5.53\% & 4.54\% \\
\textbf{Lounge} & - & 3.37\% & 3.64\% & 4.89\% & 5.22\% & 10.96\% \\
\textbf{Office} & - & 5.94\% & 5.94\% & 4.70\% & 5.17\% & 3.73\% \\
\textbf{Open Space} & - & 6.23\% & 6.41\% & 4.97\% & 5.05\% & 8.62\% \\
\textbf{Pantry} & - & 4.75\% & 4.87\% & 4.58\% & 4.78\% & 7.00\% \\
\hline
\textbf{Assembly T. 1} & - & 6.55\% & 6.56\% & 4.09\% & 3.70\% & 1.47\% \\
\textbf{Assembly T. 2} & - & 7.08\% & 7.17\% & 4.52\% & 4.09\% & 1.84\% \\
\hline
\end{tabular}
\label{tab:uncertainty-points-bayes}
\end{table}

\begin{figure}[h]
\centering
\begin{subfigure}[b]{0.85\textwidth}
   \includegraphics[width=1\linewidth]{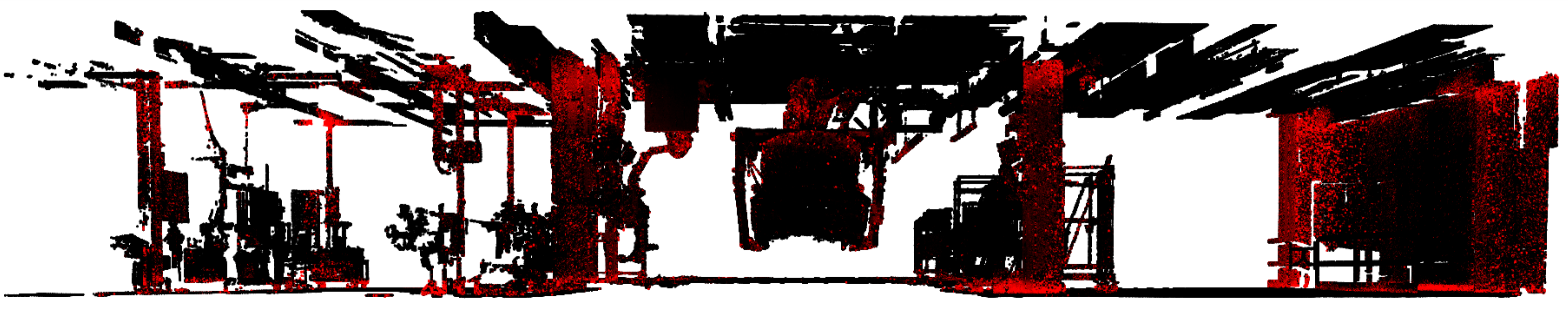}
   \caption{}
\end{subfigure}
\vspace{0.65\floatsep}
\vspace{0.1\floatsep}
\begin{subfigure}[b]{0.85\textwidth}
   \includegraphics[width=1\linewidth]{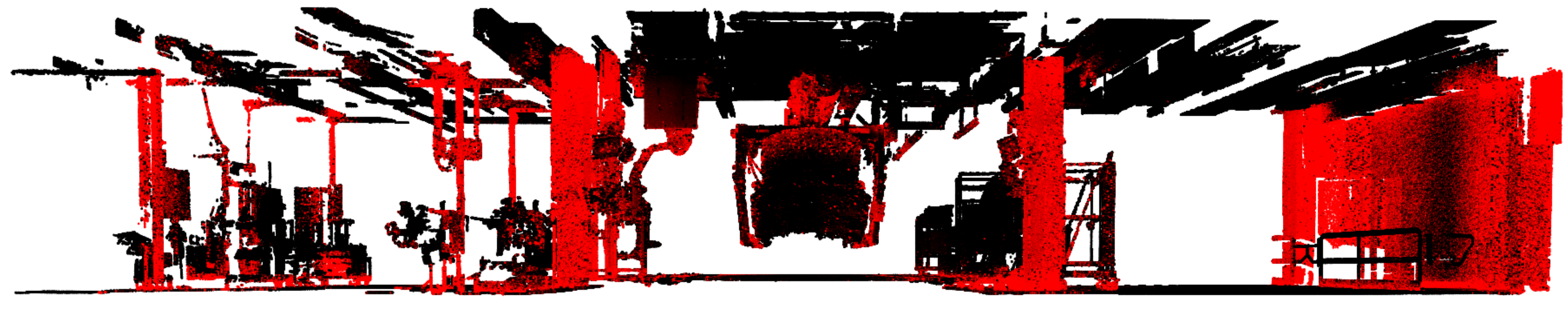}
   \caption{}
\end{subfigure}
\caption{One of the test tacts in the automotive factory data set, where certain points are illustrated in black and uncertain points are illustrated in red. The uncertainty is measured using the predictive uncertainty. \textbf{(a)}~The uncertainty threshold is set to the mean plus three sigma. \textbf{(b)}~The uncertainty threshold is set to the mean plus one sigma.}
\label{fig:test-tacts-inc-unc}
\end{figure}

Due to the choice of the uncertainty threshold about 3\% to 6\% of the predictions are dropped using predictive, aleatoric and epistemic uncertainty as well as the predictive variance. In the case of the credible interval based method considerably more or less points can be dropped as there is no explicit uncertainty threshold to be set. This leads to more accurate segmentation results, however, dropping too many predictions can cause the loss of building or facility structures.\\
In order to increase the segmentation accuracy, the methods of predictive and aleatoric uncertainty as well as the credible interval based approach are most promising. Generally, epistemic uncertainty decreases through model training, thus, the predictive uncertainty value is mainly determined by aleatoric uncertainty. This is why they lead to similar results. In most of the example cases the credible interval based uncertainty estimation method leads to the highest accuracy. Nevertheless, we will use the concept of predictive uncertainty for our upcoming observations and analyses as the credible interval based method does not allow us to control the ratio of dropped points. Further, predictive uncertainty combines the information from aleatoric and epistemic uncertainty, which is more favourable than working solely with aleatoric uncertainty.\\
Figure~\ref{fig:test-tacts-inc-unc} illustrates the importance of being able to control the ratio of dropped points. It shows one test tact of the automotive factory data set. Certain points are displayed in black and uncertain points are displayed in red. Note that uncertain points are removed in order to achieve higher segmentation accuracy. As an uncertainty measure the predictive uncertainty is used.

In Figure~\ref{fig:test-tacts-inc-unc}~(a) the uncertainty threshold is set to the mean predictive uncertainty plus three sigma.  About 3.24\% of the points are classified as uncertain and therefore displayed in red. In Figure~\ref{fig:test-tacts-inc-unc}~(b) the uncertainty threshold is the mean predictive uncertainty plus one sigma. This time about 10.69\% of the points are classified as uncertain. It can be seen that important structures like columns and walls are almost entirely classified as uncertain and thus dropped. This is in line with our expectations of Section~\ref{subsubsec:factory-data-set} as the classes of wall and column suffer from a high number of holes since they are draped with clutter objects. However, this is disadvantageous when building up a factory model as important structures are missing. Therefore, it is important to find a trade-off between achieving a higher accuracy and dropping a small number of uncertain points.

\subsubsection{Pose Estimation}
\label{subsubsec:eval-sim-gen}
After point cloud segmentation the pose estimation and simulation generation steps are carried out. For pose estimation the number of objects within each class needs to be determined as well as the points belonging to each of these objects. As we already explained in Section~\ref{subsec:workflow-pose-estimation} a clustering routine is applied to tackle these problems. Therefore, we compare the performance of different clustering algorithms including $k$-means, fuzzy $c$-means and spectral clustering. Further, we look into the density based clustering approaches DBSCAN and OPTICS. Table~\ref{tab:eval-clustering} summarizes the evaluation of the clustering algorithms for the classes of car and hanger. These are the two most difficult classes to place as they are highly asymmetric. The building parts are much easier to identify and fit. The table contains the number of points that are classified as the respective class after point cloud segmentation and down-sampling. Further, it illustrates the percentage of points that are assigned to the wrong cluster by each of the algorithms, the percentage of points where the algorithms are uncertain about the cluster assignment as well as the runtime of the clustering routines. The runtime is determined on a standard office notebook without any specific high-performance components. It can be seen that the clustering methods of $k$-means and fuzzy $c$-means are fast compared to the density based methods and spectral clustering. However, they achieve this performance only when providing them with the exact number of clusters, which makes them impractical for an automated solution. OPTICS and spectral clustering outperform DBSCAN in terms of clustering accuracy but have a higher computational time compared to DBSCAN. Thus, it is dependent on the specific use case, which of the algorithms should be applied. We decide for OPTICS as it determines the number of clusters automatically, is more accurate than DBSCAN and considerably faster than spectral clustering.

\begin{table}[h]
\centering
\caption{Evaluation of different clustering methods on points that belong to the classes of car and hanger after segmentation using the Bayesian neural network.}
\begin{tabular}{cccccc}
\hline
\textbf{Object} & \textbf{\# Points} & \textbf{Method} & \textbf{Mistakes} & \textbf{Uncertain} & \textbf{Time}\\
\hline
Car & 30 555 & $k$-means & 0.14 \% & - & 0.11 s \\
Car & 30 555 & $c$-means & 0.14 \% & 0.24 \% & 0.07 s \\
Car & 30 555 & DBSCAN & 0 \% & 1.51 \% & 3.81 s \\
Car & 30 555 & OPTICS & 0 \% & 1.51 \% & 45.64 s \\
Car & 30 555 & Spectral & 0 \% & - & 299.23 s \\
\hline
Hanger & 17 454 & $k$-means & 1.26 \% & - & 0.07 s \\
Hanger & 17 454 & $c$-means & 1.27 \% & 1.06 \% & 0.04 s \\
Hanger & 17 454 & DBSCAN & 5.17 \% & 0.91 \% & 1.74 s \\
Hanger & 17 454 & OPTICS & 0.91 \% & 5.18 \% & 15.78 s \\
Hanger & 17 454 & Spectral & 0.99 \% & - & 62.21 s \\
\hline
\end{tabular}
\label{tab:eval-clustering}
\end{table}

As already mentioned, factory layouts are often outdated especially in the case of older plants, thus, we rather compare the object positions of our generated simulation model to the labelled ground truth point cloud than to a plant layout. We place the cars and hangers of the two tacts in the test set in a simulation model based on the ground truth information and retain the gained object poses. Then we automatically generate the object poses based on the segmented point cloud using the frequentist PointNet (F), Bayesian PointNet (B) and Bayesian PointNet retaining only the predictions that are certain according to predictive uncertainty using the mean uncertainty plus two sigma as a threshold (B+U). We calculate the mean difference between the object poses gained from the ground truth and the mentioned models in terms of the x-, y- and z-coordinate as well as the roll, pitch and yaw between the two sets of object poses. As the placement for the frequentist model is very messy with this approach, we have to add additional information like a minimum height of one metre for each of the objects in order to achieve realistic placement in the scene. The Bayesian network does not need this additional information. Table~\ref{tab:eval-placement} summarizes the evaluation results of this placement.

\begin{table}[h]
\centering
\caption{Evaluation of the model placement for the classes of car and hanger. The frequentist (F), the Bayesian (B) and the Bayesian model including uncertainty (B+U) are used for segmentation. The evaluation metric is the mean deviation of the (x,y,z)-coordinates and the roll, pitch, yaw of the models generated based on the segmented point cloud and the labelled ground truth.}
\begin{tabular}{cccccccc}
\hline
\textbf{Object} & \textbf{Net} & \textbf{x-coord} & \textbf{y-coord} & \textbf{z-coord} & \textbf{Roll} & \textbf{Pitch} & \textbf{Yaw}\\
\hline
Car 1 & F & 5.48 mm & 1.13 mm & 0.45 mm & 0.19$^{\circ}$ & 0.00$^{\circ}$ & 0.06$^{\circ}$ \\
Car 2 & F & 9.09 mm & 16.55 mm & 2.74 mm & 0.49$^{\circ}$ & 0.24$^{\circ}$ & 0.31$^{\circ}$ \\
Hanger 1 & F & 3.55 mm & 0.82 mm & 6.92 mm & 0.02$^{\circ}$ & 0.15$^{\circ}$ & 0.07$^{\circ}$ \\
Hanger 2 & F & 18.03 mm & 1.41 mm & 3.92 mm & 0.26$^{\circ}$ & 0.25$^{\circ}$ & 0.01$^{\circ}$ \\
\hline
Car 1 & B & 0.61 mm & 0.34 mm & 1.17 mm & 0.13$^{\circ}$ & 0.02$^{\circ}$ & 0.03$^{\circ}$ \\
Car 2 & B & 3.49 mm & 7.69 mm & 1.39 mm & 0.17$^{\circ}$ & 0.28$^{\circ}$ & 0.22$^{\circ}$ \\
Hanger 1 & B & 7.43 mm & 6.51 mm & 3.53 mm & 0.17$^{\circ}$ & 0.28$^{\circ}$ & 0.03$^{\circ}$ \\
Hanger 2 & B & 24.58 mm & 13.17 mm & 7.01 mm & 0.40$^{\circ}$ & 0.50$^{\circ}$ & 0.16$^{\circ}$ \\
\hline
Car 1 & B+U & 1.83 mm & 6.22 mm & 0.49 mm & 0.21$^{\circ}$ & 0.07$^{\circ}$ & 0.13$^{\circ}$\\
Car 2 & B+U & 6.8 mm & 58.89 mm & 36.59 mm & 0.06$^{\circ}$ & 0.52$^{\circ}$ & 0.69$^{\circ}$\\
Hanger 1 & B+U & 2.04 mm & 0.97 mm & 2.82 mm & 0.04$^{\circ}$ & 0.06$^{\circ}$ & 0.04$^{\circ}$\\
Hanger 2 & B+U & 68.15 mm & 15.92 mm & 70.94 mm & 0.42$^{\circ}$ & 1.36$^{\circ}$ & 0.38$^{\circ}$\\
\hline
\end{tabular}
\label{tab:eval-placement}
\end{table}

\begin{figure}[h]
\centering
\begin{subfigure}[b]{0.85\textwidth}
   \includegraphics[width=1\linewidth]{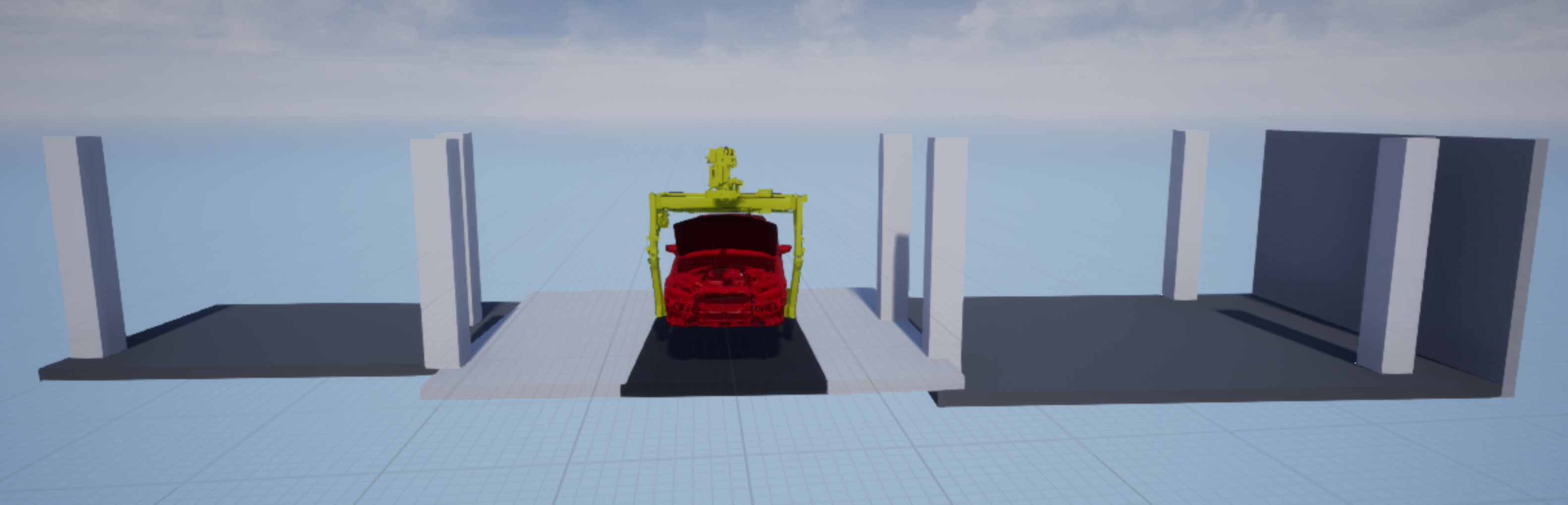}
   \caption{}
\end{subfigure}
\vspace{0.65\floatsep}
\vspace{0.1\floatsep}
\begin{subfigure}[b]{0.85\textwidth}
   \includegraphics[width=1\linewidth]{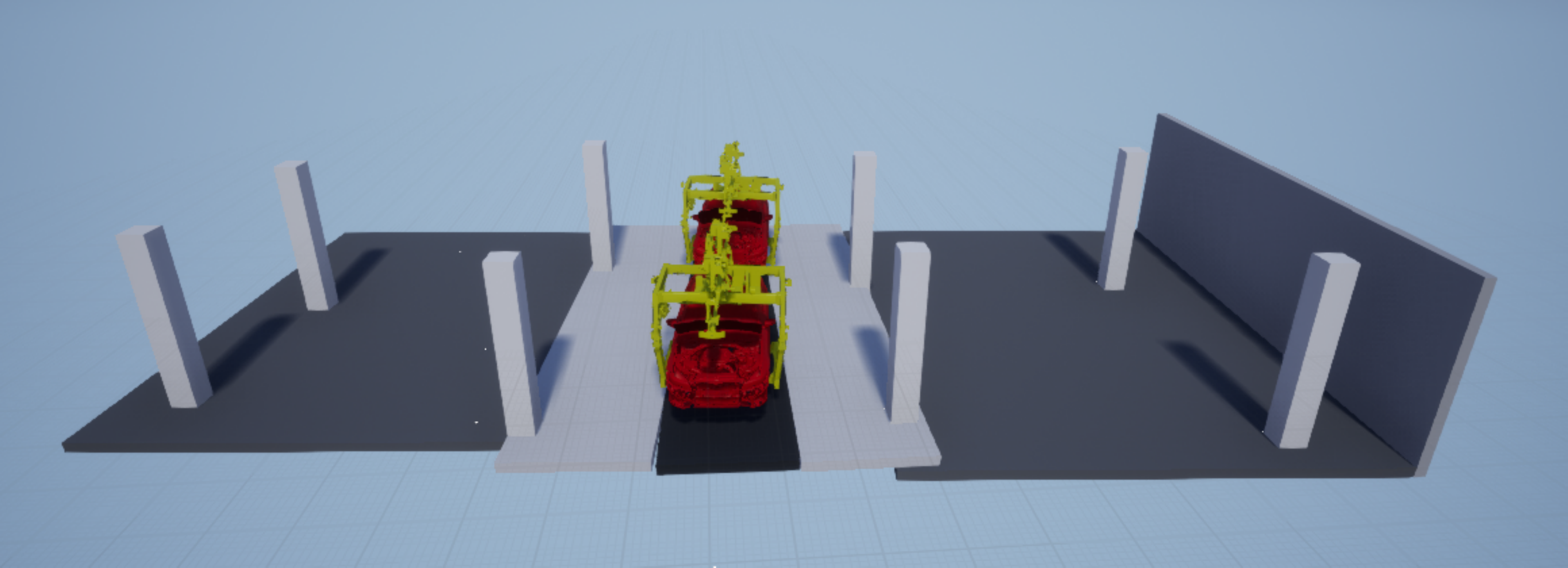}
   \caption{}
\end{subfigure}
\caption{The simulation model generated by our workflow displayed in the UE4. \textbf{(a)}~Simulation model of one test tact. \textbf{(b)} Simulation model when both test tacts are processed simultaneously.}
\label{fig:test-tacts-ue4}
\end{figure}

It has to be noted that the frequentist model performs best in terms of placement accuracy, however, it requires the input of further object information for proper model placement. In contrast to the frequentist model, the Bayesian model and the Bayesian model with additional uncertainty information achieve similar placement accuracy independent of the additional information. This happens because of the structure of the wrongly classified points in the frequentist and the Bayesian model. The frequentist model tends to misclassify bursts of points. In response to this the clustering algorithms often find more clusters than what are present in real life. The condition of objects having a minimum height of one metre discards these clusters. In contrast to that the Bayesian neural network misclassifies fewer points, which rather tend to be isolated and distributed over the scene. Therefore, the correct number of clusters is found by the clustering algorithms and an additional constraint does not change this outcome. Unexpectedly, the placement accuracy achieved by Bayesian PointNet is even better than in the case of Bayesian PointNet including the uncertainty information. Clearly, the model incorporating uncertainty achieves a higher segmentation accuracy, however, it discards about 6~\% of the points. Thus, it might happen that prominent features that are vital for object placement get lost.\\
The final simulation model for one test tact is depicted by Figure~\ref{fig:test-tacts-ue4}~(a). In this simulation model the ceiling is not displayed for clearness of illustration. Further, the class of clutter is not represented as it is irrelevant for planning tasks. Figure~\ref{fig:test-tacts-ue4}~(b) displays both test tacts processed together to generate a single simulation model.

\section{Discussion and Conclusion}
\label{sec:discussion-and-conclusion}
Aside from the presented workflow further digitalization techniques are currently being investigated by the authors to make the process more efficient. In the following we discuss the challenges and possibilities of digitalization and the generation of factory models. The economic potential of automating the digitalization process and the process of model generation is highlighted for a number of exemplary plants. Finally, a brief conclusion completes our work.

\subsection{Discussion}
In the presented workflow the digitalization step is carried out during production free times. This has the advantage of the assembly line standing still and hardly any people being in the plant except maintenance workers. Due to privacy protection regulations the faces of people that are captured by the laser scanner or the cameras have to be disguised, which is easier during production free time, as fewer people are in the building. However, personal data displayed on process boards and shift schedules still need to be handled.\\
In this work we collect our data using stationary laser scans. However, these scans usually take time. Thus, in practice it is advisable to use a combination of mobile and static laser scanners, which can still be complemented by photogrammetry if necessary. The plant is ideally digitalized using a mobile laser scanner in order to save time. Only in places where higher accuracy is needed the mobile laser scan can be complemented with a stationary one, e.g. when accurate height information plays an important role. In addition, drone technologies are investigated for digitalization purposes, however, the risk of a drone falling down and hurting employees or damaging assembled customer products has to be addressed. Further, we investigate technologies using only cameras to generate a 3D model. Generally, these methods are useful for creating a quick and rough overview of the plant. Yet, they are inadequate for most of the planning tasks due to a comparatively low accuracy.\\
Another consideration is to place laser scanners on a fleet of vehicles that are routinely used within production plants like forklifts, autonomous platforms or hangers on the production line. An incremental model of hundreds or even thousands of laser scans could be built. However, this approach causes scan activities to be carried out during the running production. On the one hand, several challenges like the registration of a high number of laser scans created by different mobile scanners, attached to different vehicle types at different heights and moving with different speeds arise. On the other hand, changes to the facility are detected quickly. Yet, not all the changes are relevant for planning projects. For instance, changes to the building structure are important for factory planning tasks in terms of spatial restrictions. In contrast to that the different parking positions of a forklift are irrelevant for building and process planning. After detection of a relevant change, a partial update of the factory point cloud is favourable, i.e. only the area where a change was detected is updated. This keeps overhead in terms of data transmission and computing power as low as possible.\\
At the moment, static simulation scenes are built from laser scans, which correspond to a snap-shot of the production plant at the time of data collection. In future, it is thinkable to generate animated simulation scenes based on a stream of input data in certain parts of the plant, i.e. process simulations are generated automatically based on periodical digitalization and data collection. For the time being, this will only be possible for smaller parts of the plant, where special sensory equipment is installed. In order to generate process simulations, the frequency of laser scans must be high. Thus, an efficient scanning strategy is a prerequisite.\\
The transition from 2D to 3D planning introduces a lot of changes in the IT and process landscape that are necessary for various planning tasks. Hence, dedicated measures of employee training need to be provided in order to support a smooth transition. Generally, the whole process from data collection over final storage to the management of access rights needs to be detailed. Further, an optimized database for point cloud storage, updating and versioning is needed. Software products that are apt for point cloud streaming could be connected to this data base as they allow for efficient visualization of the point clouds even on resource constrained devices.\\
The potential that lies in automating the factory digitalization process and model generation is huge. On the basis of quotes from three different service providers, we conclude that laser scanning of an assembly plant costs about 1-4~\euro{} per square metre depending on how much surface geometry is modelled into the scan, thus, we calculate with a low degree of modelling resulting in 1.50~\euro{} / m$^{2}$. Assuming an average area of 950,000 m$^{2}$ per production plant we calculate the potential yearly savings of $10$ plants. Further, we assume that about 60\% of the total plant area is relevant for assembly and needs to be digitalized. We aim at digitalizing a plant once a year and target a degree of automation of about 70\%. A higher degree of automation would of course be favourable, however, there are certain aspects that can hardly be automated. For instance, monitoring the digitalization process, checking on the registration targets and the quality of registration will still be handled manually.
The above-mentioned assumptions result in a yearly saving of about 6 million Euros. Table~\ref{tab:potential-auto-scan} summarizes the calculation. Note that the numbers in the above example are hypothetical and do not reflect the real volume and number of the automotive production plants under investigation in this text.
\begin{table}[h]
\centering
\caption{Potential of automating the digitalization process using laser scanners only.}
\begin{tabular}{lcc}
\hline
\textbf{Attribute} & \textbf{Value} & \textbf{Unit} \\
\hline
Costs per m$^{2}$ & 1.5 & \euro{}\\
Average area of a plant & 950,000 & m$^{2}$\\
Percentage of scanned area & 60 & \% \\
Number of plants & 10 & \# \\
Number of scans per year & 1 & \# \\
\hline
Total cost per year & 8,550,000 & \euro{} / year\\
Degree of automation & 70 & \% \\
\hline
Savings per year & 5,985,000 & \euro{} / year\\
\hline
\end{tabular}
\label{tab:potential-auto-scan}
\end{table}
~\\
Thus, the economic potential of automating the digitalization process is big. However, these are only the obvious savings gained by the automation of data collection. The availability of a comprehensive digital 3D model of big parts of the assembly area, which reflects the current state of the plant, holds an even bigger potential, when it comes to planning tasks. As already mentioned, planning mistakes can be found in the digital model rather than during the implementation of the changes. This saves a tremendous amount of money and potentially prevents production downtimes. Further, travelling efforts of planners are reduced considerably, thus, they are relieved from this additional burden and can focus on their core business. 

\subsection{Conclusion}
In this paper we describe a holistic and methodical workflow of how to generate a static simulation model in a partially automated fashion from a point cloud of a factory environment. We start with the collection of point clouds in large-scale industrial production plants using laser scanning and photogrammetry techniques. Further, the data pre-processing steps of point cloud registration, cleaning and labelling are described. The data processing steps of segmentation, pose estimation and the final simulation model generation are discussed in detail. For point cloud segmentation the application of a Bayesian neural network is evaluated as it allows the quantification of uncertainty in the network predictions. To this end, three entropy based uncertainty measures as well as the predictive variance and a credible interval based method of uncertainty estimation are discussed.\\
An automotive factory data set is collected in order to evaluate the digitalization strategy. The different technologies and hardware equipment are evaluated with respect to the resulting point cloud accuracy, completeness and point density. Based on this evaluation we conclude that laser scan point clouds are more accurate than photogrammetric point clouds but they have a lower completeness and point density. Further, the use of wide-angle instead of fish-eye lenses for photogrammetry is highly recommended as the former outperforms the latter with respect to all the evaluation metrics. Thus, in terms of data collection we recommend to combine the point clouds generated by laser scanners and photogrammetry techniques.\\
The subsequent steps of the proposed workflow are evaluated on this data set as well. The segmentation step is additionally evaluated on the publicly available S3DIS data set. The Bayesian segmentation network clearly outperforms the frequentist model even without considering the additional information provided by the uncertainty in the network predictions. For a more detailed evaluation of the Bayesian segmentation network see~\citep{petschnigg2020uncertainty}. Taking into account the network's uncertainty in its predictions and evaluating the segmentation accuracy only on certain points increases the network performance considerably using any of the described uncertainty measures. In this case uncertain predictions are discarded. The notions of predictive and aleatoric uncertainty as well as the credible interval based method provide us with the best results. In the given use case, we decide to focus on the measure of predictive uncertainty as it allows us to control the amount of points dropped by setting an uncertainty threshold, which is not the case for the credible interval based method. As predictive uncertainty combines the information of both aleatoric and epistemic uncertainty, we prefer this measure over aleatoric uncertainty alone. Based on the segmented point cloud we estimate the poses of relevant objects using, among other methods, clustering techniques. Therefore, the common clustering algorithms $k$-means and fuzzy $c$-means, the density based methods DBSCAN and OPTICS as well as spectral clustering are compared. Given that $k$-means and $c$-means clustering achieve superior performance when knowing the number of clusters in advance we advise not to use these methods for an automated solution. All the remaining methods automatically determine the number of clusters. The OPTICS algorithm provides the best trade-off between runtime and clustering accuracy. Thus, we apply this algorithm for the final evaluation of the placement accuracy. The object placement is carried out on the basis of the segmented point cloud using the frequentist and the Bayesian neural network. Additionally, the Bayesian neural network including uncertainty information is used for object placement. The placement in the frequentist case is by far the worst, not producing any reasonable simulation scene without adding additional hand-crafted constraints for object placement. Astonishingly, the placement accuracy is highest for the Bayesian neural network not considering network uncertainty, even though the segmentation accuracy is considerably improved by incorporating uncertainty information. Most probably, the reason is that by considering network uncertainty relevant features of the point cloud are classified as uncertain and thus are dropped. Therefore, we conclude that the Bayesian neural network without removing uncertain points is best suited for our use case.


\authorcontributions{Conceptualization, C.P.; methodology, C.P.; software, C.P., M.S. and L.W.; validation, C.P., M.S. and L.W.; formal analysis, C.P. and J.P.; investigation, C.P.; resources, C.P.; data curation, C.P. and M.S.; writing---original draft preparation, C.P., M.S. and J.P.; writing---review and editing, C.P. and J.P.; visualization, C.P., M.S. and L.W.; supervision, J.P.; project administration, C.P. and J.P. All authors have read and agreed to the published version of the manuscript.}

\funding{This research received no external funding.}




\conflictsofinterest{The authors declare no conflict of interest.} 



\abbreviations{The following abbreviations are used in this manuscript:\\

\noindent 
\begin{tabular}{@{}ll}
2D & two-dimensional\\
3D & three-dimensional\\
AML & Automation Markup Language\\
BNN & Bayesian Neural Network\\
CAD & Computer Aided Design\\
DBSCAN & Density-Based Spatial Clustering of Applications with Noise\\
DoF & Degrees of Freedom\\
ELBO & Evidence Lower Bound\\
ICP & Iterative Closest Points\\
IoU & Intersection over Union\\
KL & Kullback-Leibler\\
OEM & Original Equipment Manufacturer\\
OPTICS & Ordering Points To Identify the Clustering Structure\\
RANSAC & Random Sample Consensus\\
S3DIS & Stanford Large-Scale 3D Indoor Spaces Data Set\\
UE4 & Unreal Engine 4\\
VDI & Verein Deutscher Ingenieure (Association of German Engineers)

\end{tabular}}

\reftitle{References}

\externalbibliography{yes}
\bibliography{references}

\end{document}